\documentclass[10pt,twocolumn,letterpaper]{article}

\usepackage{iccv}
\usepackage{times}
\usepackage{epsfig}
\usepackage{graphicx}
\usepackage{amsmath}
\usepackage{amssymb}

\usepackage{blindtext}
\usepackage{graphicx}
\usepackage{amssymb}
\usepackage{graphicx}
\usepackage{bbm}
\usepackage{mathtools}
\usepackage{mathrsfs}
\usepackage{multirow}
\usepackage{algorithm}
\usepackage{algorithmic}
\usepackage{bbding}
\usepackage{epsfig}
\usepackage{amsmath}
\usepackage{subfigure}
\usepackage{times}
\usepackage{epsfig}
\usepackage{graphicx}
\usepackage{amsmath}
\usepackage{amssymb}
\usepackage{bm}
\usepackage{amsfonts,amssymb}
\usepackage{booktabs}
\usepackage{float}
\usepackage{subfigure}
\usepackage{makecell}
\usepackage{multirow}

\usepackage[breaklinks=true,
            bookmarks=false,
            colorlinks,
            linkcolor=red,       
            anchorcolor=blue,  
            citecolor=green, 
            ]{hyperref}

\iccvfinalcopy 

\def\iccvPaperID{****} 

\ificcvfinal\fi

\begin{document}

\title{Dynamic Attentive Graph Learning for Image Restoration}


\author{
Chong Mou$^{\dagger}$, Jian Zhang$^{\dagger, \ddagger}$, Zhuoyuan Wu$^{\dagger}$\\
$^\dagger$Peking University Shenzhen Graduate School, Shenzhen, China\\
$^\ddagger$Peng Cheng Laboratory, Shenzhen, China\\
{\tt\small eechongm@gmail.com; zhangjian.sz@pku.edu.cn; wuzhuoyuan@pku.edu.cn}
}

\maketitle
\let\thefootnote\relax\footnotetext{This work was supported in part by National Natural Science Foundation of China (61902009). (\textit{Corresponding author: Jian Zhang.}) }
\ificcvfinal\thispagestyle{empty}\fi



\begin{abstract}
Non-local self-similarity in natural images has been verified to be an effective prior for image restoration. However, most existing deep non-local methods assign a fixed number of neighbors for each query item, neglecting the dynamics of non-local correlations. Moreover, the non-local correlations are usually based on pixels, prone to be biased due to image degradation. To rectify these weaknesses, in this paper, we propose a dynamic attentive graph learning model (DAGL) to explore the dynamic non-local property on patch level for image restoration. Specifically, we propose an improved graph model to perform patch-wise graph convolution with a dynamic and adaptive number of neighbors for each node. In this way, image content can adaptively balance over-smooth and over-sharp artifacts through the number of its connected neighbors, and the patch-wise non-local correlations can enhance the message passing process. Experimental results on various image restoration tasks: synthetic image denoising, real image denoising, image demosaicing, and compression artifact reduction show that our DAGL can produce state-of-the-art results with superior accuracy and visual quality. The source code is available at \href{https://github.com/jianzhangcs/DAGL}{https://github.com/jianzhangcs/DAGL}.
\end{abstract}

\section{Introduction}
Image restoration (IR) is typically an ill-posed inverse problem aiming to restore a high-quality image ($\mathbf{I}_{HQ}$) from its degraded measurement ($\mathbf{I}_{LQ}$) corrupted by various degradation factors. The degradation process can be defined as $\mathbf{I}_{LQ}=\mathbf{H}\mathbf{I}_{HQ}+\mathbf{n}$, where $\mathbf{H}$ is a linear degradation matrix, and $\mathbf{n}$ represents additive noise \cite{zhang2014csvt,zhao2016reducing}. According to $\mathbf{H}$, IR can be categorized into many subtasks, \textit{e.g.}, denoising, compression artifact reduction, demosaicing, super-resolution, compressive sensing \cite{dncnn, rcan, zhao2016video, zhang2020optimization}.

The rise of deep learning has greatly facilitated the development of image restoration. Many deep learning-based methods \cite{dncnn,ircnn,ffdnet,memnet} have been proposed to solve this ill-posed problem. Despite the remarkable success, most methods focus on learning from a lot of external training data without fully utilizing the internal prior in images. By contrast, many classic model-based methods are implemented based on various priors, \textit{e.g.}, total variation~\cite{tv}, sparse representation~\cite{sp1,sp2,GSR2014}, and self-similarity~\cite{nonlocalmean,BM3D}. The self-similarity assumes that similar content would recur across the whole image, and the local content can be recovered with the help of similar items from other places. Inspired by~\cite{nonlocalmean}, non-local neural networks \cite{nonlocalnet} utilized self-similarity via deep networks, which are subsequently introduced to many image restoration tasks \cite{nlrn,rnan}. However, these pixel-wise non-local methods are easily influenced by noisy signals within corrupted images. \cite{nlnet,unlnet} were proposed to establish long-range correlations on patch level. Nevertheless, the patch matching step is isolated from the training process. In N3Net~\cite{n3net}, a differentiable $K$-Nearest Neighbor (KNN) method was proposed. However, restricted by the high complexity of channel-wise feature fusion, N3Net can only perform the non-local operation within a small search region ($10\times 10$) and a small number of matching patches. Some very recent methods~\cite{cola,nl2020,ipt} proposed more efficient patch-wise non-local methods. But they followed the same paradigm as existing non-local methods to construct fully connected correlations. 

In general, the repeatability of different image content is distinct, causing different requirements of non-local correlations in restoring different image content. An early work~\cite{dynon} has well studied this property, finding that smooth image contents recur much more frequently than complex image details, and they should be treated differently. 

Graph convolutional network (GCN) is a special non-local method designed to process the graph data by establishing long-range correlations in non-Euclidean space. However, the large domain gap limits the application of this flexible non-local method in computer vision community. Recently, few works \cite{gcdn,gcdn_b,face_gcn} proposed to apply GCN to image restoration tasks. Specifically, \cite{gcdn} and \cite{gcdn_b} are built based on Edge-Conditioned Convolution (ECC) \cite{ecc} for image denoising. However, they constructed the long-range correlations based on pixels and assigned a fixed number of neighbors for each graph node. In~\cite{face_gcn}, a patch-wise GCN method is proposed for facial expression restoration. Nevertheless, the adjacency matrix is predefined based on the facial structure and isolated from the training process. In addition to ECC, graph attention network (GAT)~\cite{gat} is a popular graph model combined with attention mechanism to identify the importance of different neighboring nodes. 

Inspired by GAT, in this paper, we propose a novel dynamic attentive graph learning model (DAGL) for image restoration. In our proposed DAGL, the corrupted image is recovered in an image-specific and adaptive graph constructed based on local feature patches. 

\section{Related Works}
Our model is closely related to image restoration algorithms, non-local attention methods, and graph convolutional networks. Since in what follows, we give a brief review of these aspects and some most relevant methods.

\subsection{Image Restoration Architectures}
Driven by the success of deep learning, almost all recent top-performing image restoration methods are implemented based on deep networks. Stacking convolutional layers is the most well-known CNN-based strategy. Dong \textit{et al}. proposed ARCNN \cite{arcnn} for image restoration with several stacked convolutional layers. Subsequently,~\cite{dncnn,ffdnet,ircnn} utilized deeper convolutional architecture and residual learning to further enhance image restoration performance. Recently, abundant novel models and function units were proposed. MemNet \cite{memnet} utilized the dense connection in convolutional layers for image denoising. To enlarge the receptive field, hourglass-shaped architecture \cite{cbdnet,vdnet,unprocessingnet,gdanet,aindnet,deamnet}, dilated convolution \cite{ircnn,dilated}, and very deep residual networks~\cite{rcan,rnan} are often used. However, most methods are plain networks and neglect to use non-local information.

\subsection{Non-local Prior for Image Restoration}
Non-local self-similarity is an effective prior that has been widely used in image restoration tasks. Some classic methods \cite{BM3D,nonlocalmean} utilized self-similarity for image denoising and achieved attractive performance. Following the importance of self-similarity, some recent approaches \cite{rnan,nlrn} utilized this prior based on non-local neural networks \cite{nonlocalnet}. Moreover, some patch-wise non-local methods \cite{nlnet,unlnet,n3net} or transformer-based methods~\cite{ipt,cola} were proposed. These methods performed matching and aggregation in a non-local manner can be generally defined as:
\begin{equation}
    \hat{\mathbf{x}}_{i} = \frac{1}{z_{i}}\sum_{{j}\in \mathbb{Q}}\phi(\mathbf{y}_{i},\mathbf{y}_{j})G(\mathbf{y}_{j}), \forall i,
\label{eq1}
\end{equation}
where $\mathbb{Q}$ refers to the search region, and $z_{i}$ represents the normalizing constant calculated by $z_{i}=\sum_{{j}\in \mathbb{Q}}\phi(\mathbf{y}_{i},\mathbf{y}_{j})$. The function $\phi$ computes pair-wise affinity between query item $\mathbf{y}_{i}$ and key item $\mathbf{y}_{j}$. $G$ is a feature transformation function that generates a new representation of $\mathbf{y}_{j}$. While the above operation aggregates adequate information for the query item, the feature aggregation is restricted to be fully connected, involving all features within the search region, no matter how similar they are to the query item.
\begin{figure*}[t]
\small
    \centering
    \includegraphics[width=0.95\linewidth,height=4.5cm]{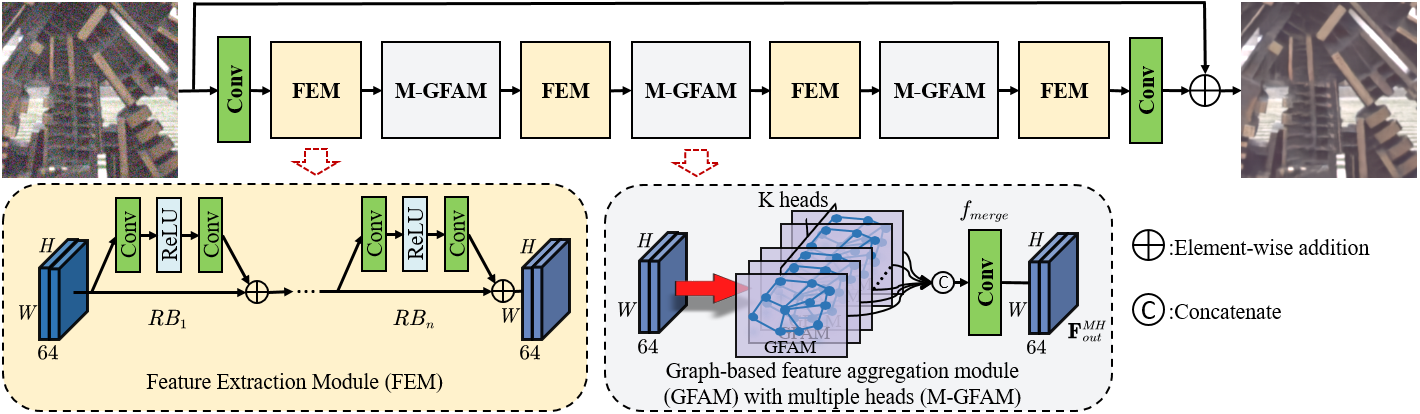}
    \caption{Proposed dynamic attentive graph learning model (DAGL). The feature extraction module (FEM) employs residual blocks to~extract deep features. The graph-based feature aggregation module (GFAM) constructs a graph with dynamic connections and performs~patch-wise graph convolution. GFAM with multiple-heads (M-GFAM) jointly aggregates information from different representation~subspaces.}
    \label{network}
    \vspace{-8pt}
\end{figure*}
\subsection{Graph Convolutional Networks (GCN)}
By extending convolutional neural networks (CNN) from grid data, such as images and videos, to graph-structured data, GCN has been attracting growing attention from the computer vision community due to its robust capacity of non-local feature aggregation. Not that without loss of generality, the commonly used non-local neural networks \cite{nonlocalnet} can be viewed as a fully connected graph~\cite{fcgcn}. Recently, \cite{face_gcn} utilized the predefined adjacency matrix to perform graph convolution for facial expression restoration. \cite{gcdn,gcdn_b} applied Edge-Conditioned Convolution (ECC)~\cite{ecc}, a well-known GCN method, to image denoising task. \cite{ecc3d} further extended ECC to 3D denoising tasks. Let us consider a graph that contains $N$ nodes: $\mathcal{G=(V,E)}$, where $\mathcal{V}$ is the set of graph nodes, and $\mathcal{E}$ is the set of edges. Let $\mathbf{v}_{i}\in \mathbb{R}^{l_1}$ denote a graph node and $\mathbf{e}_{ij}=(\mathbf{v}_i, \mathbf{v}_j)\in \mathcal{E}$ denote an edge pointing from $\mathbf{v}_j$ to $\mathbf{v}_{i}$. In ECC, there exists a shared filter-generating network $\mathcal{F}$: $\mathbb{R}^s \mapsto \mathbb{R}^{l_2\times l_1}$. Given an edge label $\mathbf{e}_{ij}\in \mathbb{R}^s$, it outputs an edge-specific embedding matrix $\mathbf{\Theta}_{ij} \in \mathbb{R}^{l_2\times l_1}$. The aggregation process of ECC is an averaging operation embedded by the edge-specific embedding matrix, which can be formalized as:
\begin{equation}
    \hat{\mathbf{v}}_{i}=\frac{1}{|\mathcal{N}_{i}|}\sum_{j\in \mathcal{N}_{i}}\mathcal{F}(\mathbf{e}_{ij})\mathbf{v}_{j}+\mathbf{b}=\frac{1}{|\mathcal{N}_{i}|}\sum_{j\in \mathcal{N}_{i}}\mathbf{\Theta}_{ij}\mathbf{v}_{j}+\mathbf{b},
\end{equation}
where $\mathcal{N}_{i}=\{j|(\mathbf{v}_i, \mathbf{v}_j)\in \mathcal{E}\}$ is the set of indexes of neighboring nodes of $\mathbf{v}_i$, and $\mathbf{b}\in \mathbb{R}^{l_2}$ is a learnable bias. Apart from ECC, graph attention network (GAT) \cite{gat} is also a popular GCN method, and our proposed DAGL is inspired by this method. Unlike ECC generating an embedding matrix through the edge label to perform embedding and averaging aggregation, GAT developed an attention weight for each edge based on the self-attention mechanism \cite{atisal}. In this way, each node can aggregate the information selectively from all its connected neighbors. The calculation of the attention weight is defined as:
\begin{equation}
 \alpha_{ij} = \frac{exp(LeakyReLU(\mathbf{a}^{T}[\mathbf{W}\mathbf{v}_{i}||\mathbf{W}\mathbf{v}_{j}]))}{\sum_{k\in \mathcal{N}_{i}}exp(LeakyReLU(\mathbf{a}^{T}[\mathbf{W}\mathbf{v}_{i}||\mathbf{W}\mathbf{v}_{k}]))},
\label{gat}
\end{equation}
where $\mathbf{W}\in \mathbb{R}^{l_2\times l_1}$ and $\mathbf{a}\in \mathbb{R}^{2l_2\times 1}$ refer to the learnable weight matrixes of shared linear transformations, and $||$ represents the concatenating operation. In the process of aggregation, the source node will be updated through the sum of all its connected neighbors weighted by the learnable attention weights:
\begin{equation}
    \hat{\mathbf{v}}_{i} = \sum_{j\in \mathcal{N}_{i}}\alpha_{ij}\cdot \mathbf{W}\mathbf{v}_{j}.
    \label{sum}
\end{equation}
Different from most GCN methods directly processing graph data, the main challenge of applying GCN to the image restoration community is how to construct a graph and perform graph convolution on regular grid data effectively. In this paper, we propose an improved graph attention model to perform patch-wise graph convolution with dynamic graph connections for image restoration. The proposed method achieves state-of-the-art performance on various image restoration tasks. 
\begin{figure*}[t]
\small
    \centering
    \includegraphics[width=.98\linewidth]{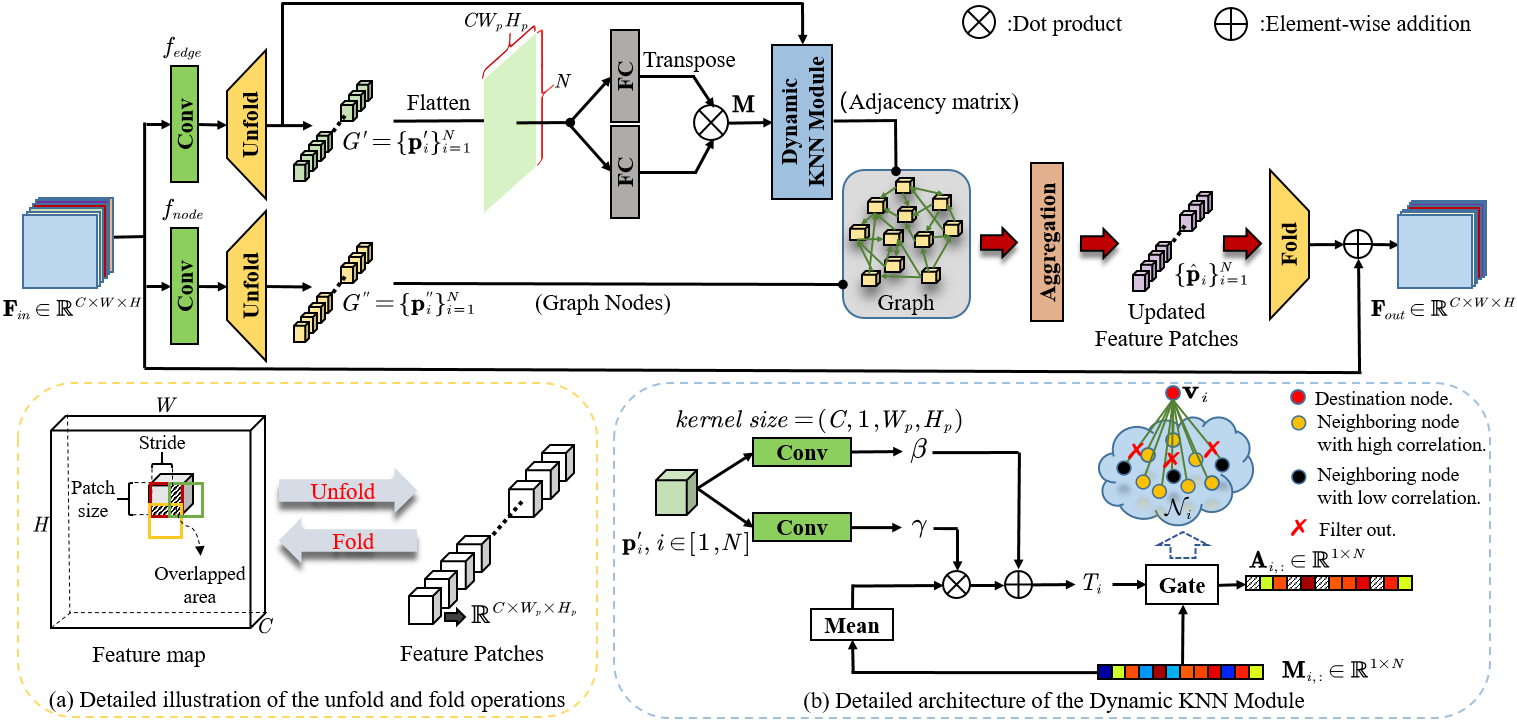}
    \caption{Detailed illustration of the proposed graph-based feature aggregation module (GFAM). The subfigure (a) elaborates the unfold and fold operations. The subfigure (b) presents the detailed architecture of the dynamic KNN module, which is used to generate a node-specific threshold to filter out graph connections with low importance.}
    \label{graph}
    \vspace{-8pt}
\end{figure*}
\section{Proposed Method}

\subsection{Framework}
An overview of our proposed model (DAGL) is shown in Fig.~\ref{network}, mainly composed of two components: feature extraction module (FEM) and graph-based feature aggregation module (GFAM) with multiple heads (M-GFAM). Similar to many image restoration networks, we add a global pathway from the input to the final output, which encourages the network to bypass low-frequency information. The feature extraction module comprises several residual blocks (RBs), and we follow the strategy in~\cite{rnan} to remove batch normalization \cite{bn} layers from residual blocks. The graph-based feature aggregation module is the core of our proposed DAGL, which is implemented based on graph attention networks (GAT) \cite{gat}. More details about GFAM will be given in the following subsection. 

Our proposed model is optimized with the $\mathcal{L}2$ loss function. Given a training set $\{\mathbf{I}_{HQ}^{b},\mathbf{I}_{LQ}^{b}\}_{b=1}^{B}$, which contains $B$ training pairs. The goal of the training can be defined as:
\begin{equation}
    \mathcal{L}(\theta )=\frac{1}{B}\sum_{b=1}^{B}\left\|\mathbf{I}_{HQ}^{b}-\mathcal{H}(\mathbf{I}_{LQ}^{b})\right\|^2_{2},
\end{equation}
where $\mathcal{H}$ refers to the function of our proposed DAGL, and $\theta$ refers to the learnable parameters.

\subsection{Graph-based Feature Aggregation Module}
As mentioned previously, existing deep non-local methods and graph-based image restoration methods assigned a fixed number of neighbors for each query/node. The main difference is that deep non-local methods involve all items in the search region to update one query item, and the graph-based methods select $K$ nearest neighbors for each node. In this subsection, we will present our proposed graph-based feature aggregation module (GFAM), a dynamic solution to break this set pattern. Our GFAM constructs the long-range correlations based on 3D feature patches and assigns a dynamic number of neighbors for each query patch. The detailed illustration of our proposed GFAM is shown in Fig.~\ref{graph}, which is mainly composed of three phases: initialization, dynamic graph construction, and feature aggregation.  

\textbf{Initialization.} In our GFAM, we first need to initialize some elements for constructing a graph $\mathcal{G}=(\mathcal{V},\mathcal{E})$ on regular grid data, where $\mathcal{V}$ is the set of nodes with $|\mathcal{V}|=N$ and $\mathcal{E}\subseteq \mathcal{V}\times \mathcal{V}$ is the set of edges. Assuming $N$ overlapped feature patches $\{\mathbf{p}_{i}\}_{i=1}^{N}$ with the patch size being $C\times W_p \times H_p$ ($C\times 7\times 7$ by default), neatly arranged in the input feature map $\mathbf{F}_{in}\in \mathbb{R}^{C\times W\times H}$. We apply two $1\times 1$ convolutional layers ($f_{edge}$ and $f_{node}$) to transform $\mathbf{F}_{in}$ to two independent representations and then utilize the unfold operation to extract the transformed feature patches into two groups: $G^{\prime }=\{\mathbf{p}_i^{\prime}\}_{i=1}^{N}$ and $G^{\prime \prime }=\{\mathbf{p}_i^{\prime \prime}\}_{i=1}^{N}$. The feature patches in $G^{\prime}$ and $G^{\prime \prime }$ have the following feature representations:
\begin{equation}
   \begin{cases}
    & \mathbf{p}_i^{\prime} = f_{edge}(\mathbf{p}_i)\\
    & \mathbf{p}_i^{\prime \prime} = f_{node}(\mathbf{p}_i).
    \end{cases}
\end{equation}
$G^{\prime }$ is used to build graph connections ($\mathcal{E}$), and $G^{\prime \prime}$ is assigned as the graph nodes.

\textbf{Dynamic Graph Construction.} The graph nodes in our method are directly assigned by feature patches in $G^{\prime \prime}$: $\mathcal{V}=G^{\prime \prime}$. 
In establishing graph connections, we select a dynamic number of neighbors for each node based on the nearest principle. For this purpose, we design a dynamic KNN module to generate an adaptive threshold for each node to select neighbors whose similarities are above the threshold. Specifically, given the set of feature patches $G^{\prime }$, we first flatten each feature patch into a feature vector. The pair-wise similarities can be efficiently calculated by dot product, producing a similarity matrix $\mathbf{M}\in \mathbb{R}^{N\times N}$. Let us consider $\mathbf{M}_{i,:}$, the $i$-th row of $\mathbf{M}$, representing similarities between the $i$-th node and the other nodes. The average of $\mathbf{M}_{i,:}$ is the fairly average importance of different nodes to the $i$-th node. Thus, it is an appropriate choice for the threshold, represented as $T_i$. As illustrated in Fig.~\ref{graph}(b), to improve the adaptability, we further apply a node-specific affine transformation to compute $T_i$:
\begin{equation}
    T_i = \frac{\psi_{1}(\mathbf{p}_i^{\prime})}{N} \sum_{k=1}^{N}\mathbf{M}_{i,k} + \psi_{2}(\mathbf{p}_i^{\prime}) = \frac{\gamma}{N} \sum_{k=1}^{N}\mathbf{M}_{i,k} + \beta,
\end{equation}
where $\psi_{1}$ and $\psi_{2}$ are two independent convolutional layers with the kernel size being $C\times 1\times W_{p} \times H_{p}$ to embed each node to specific affine transformation parameters ($\beta$, $\gamma$). To achieve a differentiable threshold truncation, we utilize ReLU~\cite{relu} function to truncate the negative part and keep the positive part. This process is formalized as: 
\begin{equation}
    \mathbf{A}_{i,:} = ReLU(\mathbf{M}_{i,:}-T_{i}),
\end{equation}
where $\mathbf{A}\in \mathbb{R}^{N\times N}$ is the adjacency matrix in which $\mathbf{A}_{ij}$ is assigned the similarity weight if $\mathbf{p}_j^{\prime}$ connects to $\mathbf{p}_i^{\prime}$, otherwise equal to zero.  
Next, following the definition in Eq.~\ref{gat}, we normalize the similarity of all connected nodes (non-zero values in $\mathbf{A}_{i,:}$) by the softmax function to calculate the attention weights:
\begin{equation}
    \alpha_{ij}=\frac{exp(\mathbf{A}_{ij})}{\sum_{k\in \mathcal{N}_{i}}exp(\mathbf{A}_{ik})}, j\in \mathcal{N}_i.
\end{equation}

\begin{figure*}[t]
\scriptsize
    \centering
    \begin{minipage}[t]{0.12\linewidth}
    \centering
        \includegraphics[width=1\columnwidth,height=2cm]{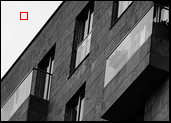}\\
        \includegraphics[width=1\columnwidth,height=2cm]{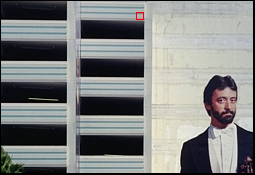}\\Query patch in high-quality label
    \end{minipage}
        \begin{minipage}[t]{0.12\linewidth}
            \centering
        \includegraphics[width=1\columnwidth,height=2cm]{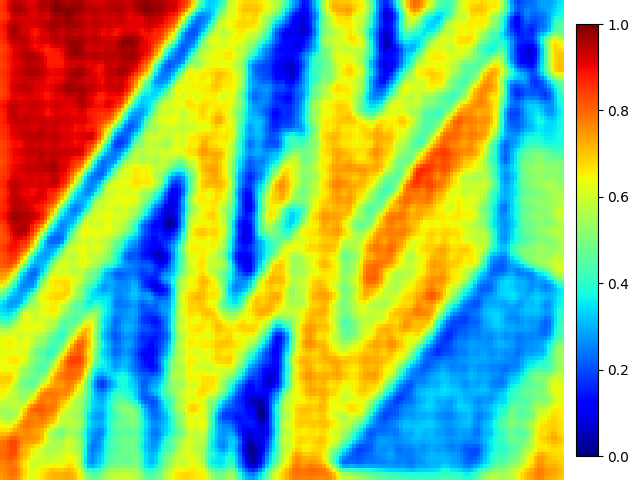}\\
        \includegraphics[width=1\columnwidth,height=2cm]{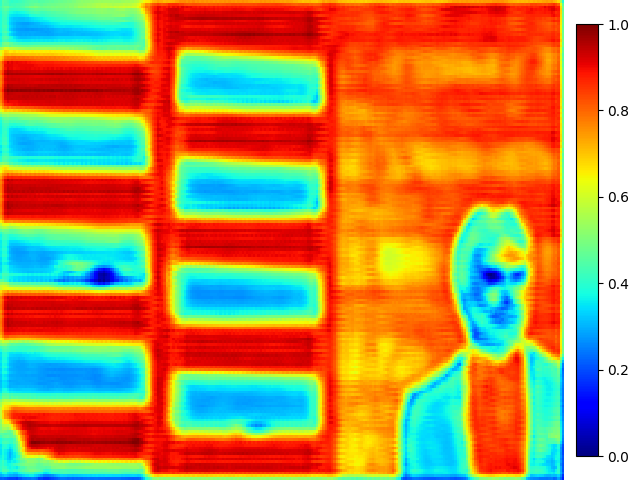}\\Similarity matrix
    \end{minipage}
        \begin{minipage}[t]{0.12\linewidth}
        \centering
        \includegraphics[width=1\columnwidth,height=2cm]{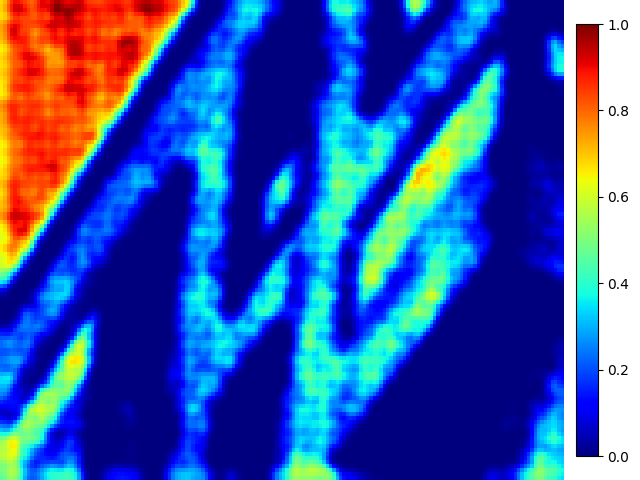}\\
        \includegraphics[width=1\columnwidth,height=2cm]{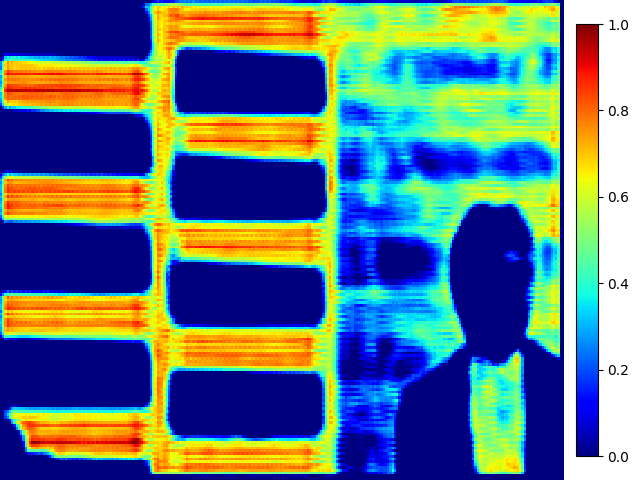}\\adjacency matrix
    \end{minipage}
        \begin{minipage}[t]{0.12\linewidth}
        \centering
        \includegraphics[width=1\columnwidth,height=2cm]{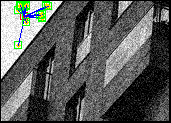}\\
        \includegraphics[width=1\columnwidth,height=2cm]{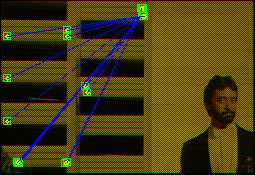}\\Graph connection samples
    \end{minipage}
        \begin{minipage}[t]{0.12\linewidth}
    \centering
        \includegraphics[width=1\columnwidth,height=2cm]{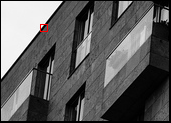}\\
        \includegraphics[width=1\columnwidth,height=2cm]{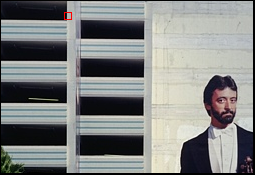}\\Query patch in high-quality label
    \end{minipage}
        \begin{minipage}[t]{0.12\linewidth}
            \centering
        \includegraphics[width=1\columnwidth,height=2cm]{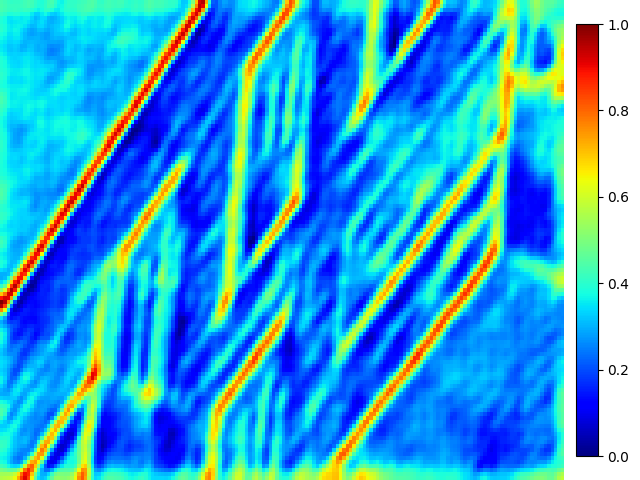}\\
        \includegraphics[width=1\columnwidth,height=2cm]{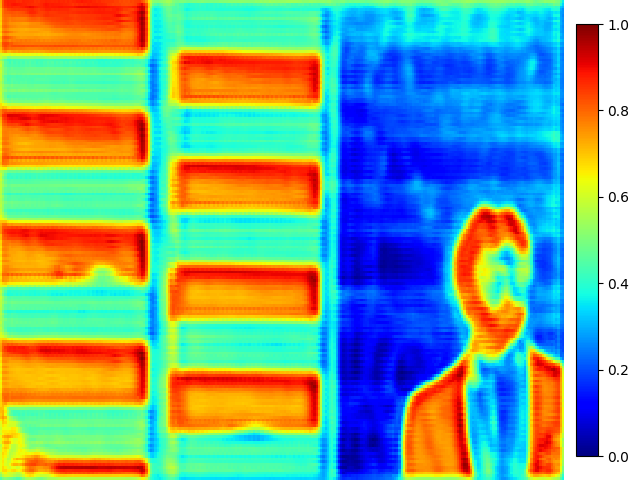}\\Similarity matrix
    \end{minipage}
        \begin{minipage}[t]{0.12\linewidth}
        \centering
        \includegraphics[width=1\columnwidth,height=2cm]{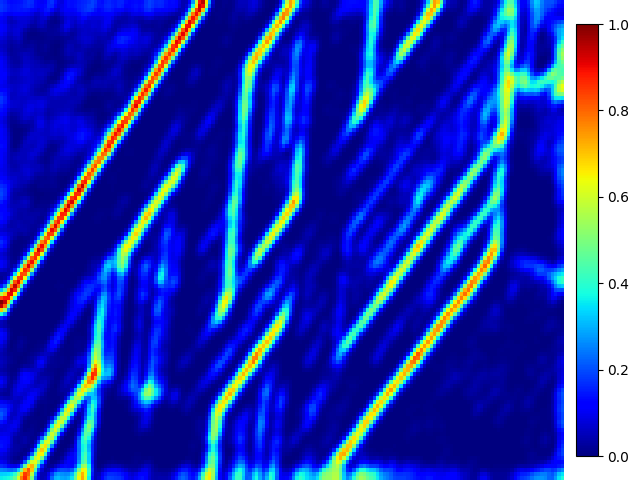}\\
        \includegraphics[width=1\columnwidth,height=2cm]{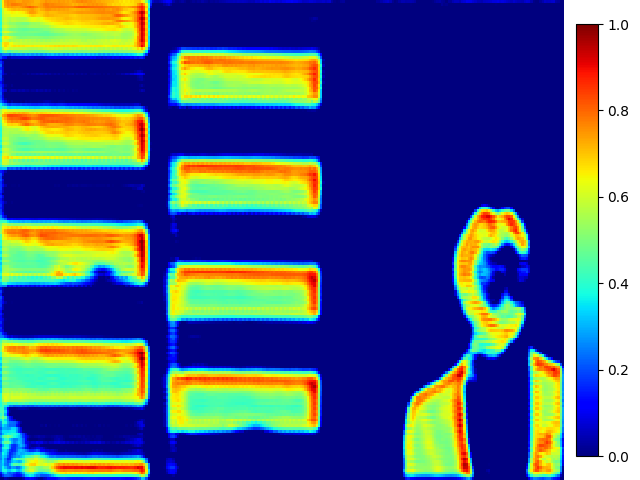}\\adjacency matrix
    \end{minipage}
        \begin{minipage}[t]{0.12\linewidth}
        \centering
        \includegraphics[width=1\columnwidth,height=2cm]{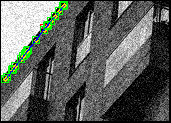}\\
        \includegraphics[width=1\columnwidth,height=2cm]{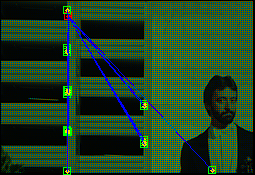}\\Graph connection samples
    \end{minipage}
    \vspace{2pt}
    \caption{Visualization of construction of graph connections. The regions of query patches are labeled with red boxes. The similarity matrixes and adjacency matrixes are presented in the form of heat maps. The lighter color indicates higher similarity/importance. For illustration purpose, we present some highly correlated neighbors (labeled with green boxes). One can see that our method can capture satisfied long-range correlations in highly degraded images, and the adjacency matrix accurately filters out low-important correlations.}
    \vspace{-8pt}
    \label{fig:ht_graph}
\end{figure*}

\begin{figure}[t]
    \centering
    \begin{minipage}[t]{0.32\linewidth}
    \centering
        \includegraphics[width=1\columnwidth,height=2.5cm]{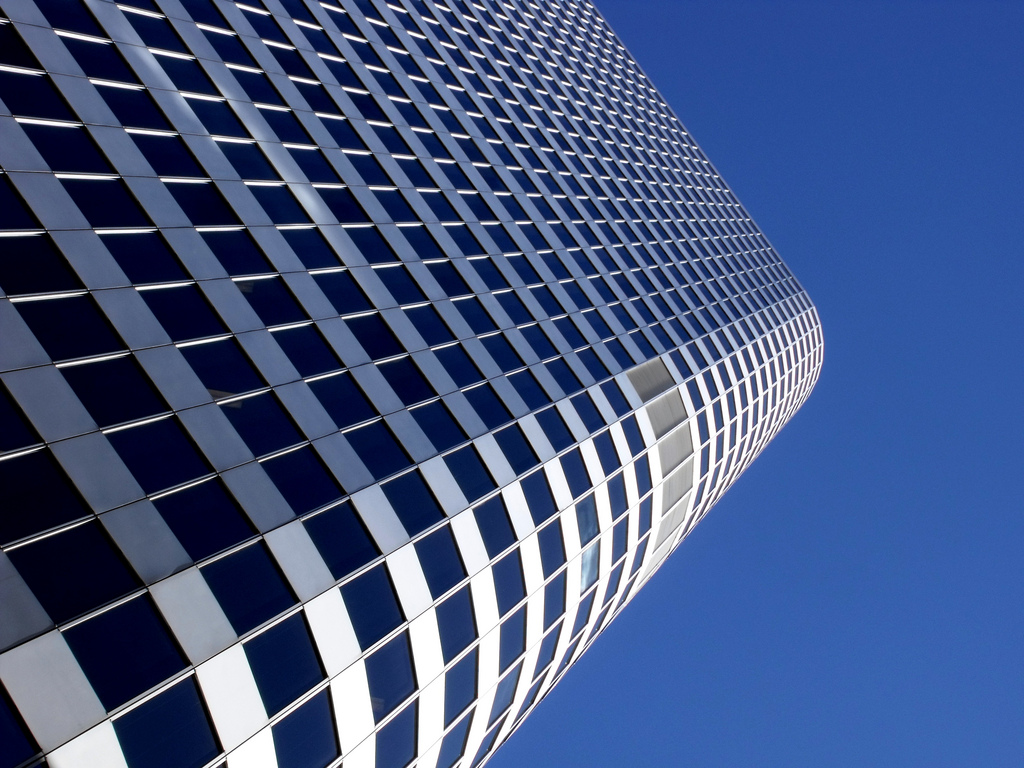}\\
        \includegraphics[width=1\columnwidth,height=2.5cm]{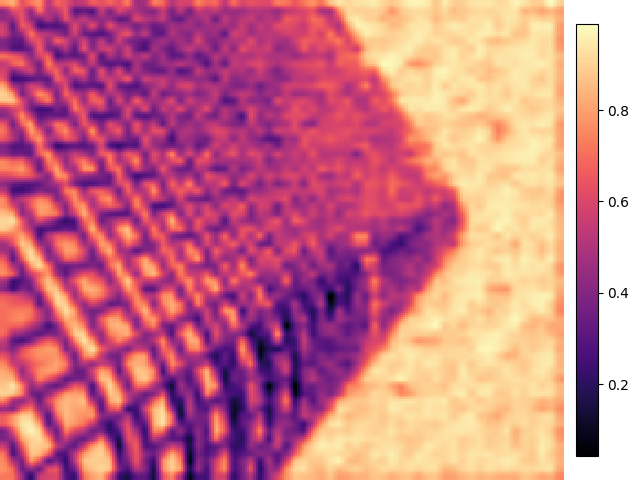}
    \end{minipage}
        \begin{minipage}[t]{0.32\linewidth}
            \centering
        \includegraphics[width=1\columnwidth,height=2.5cm]{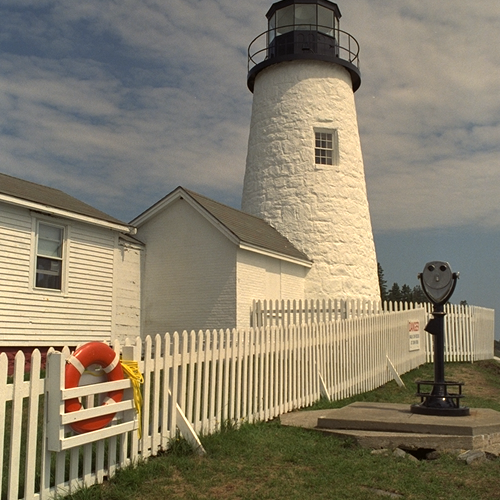}\\
        \includegraphics[width=1\columnwidth,height=2.5cm]{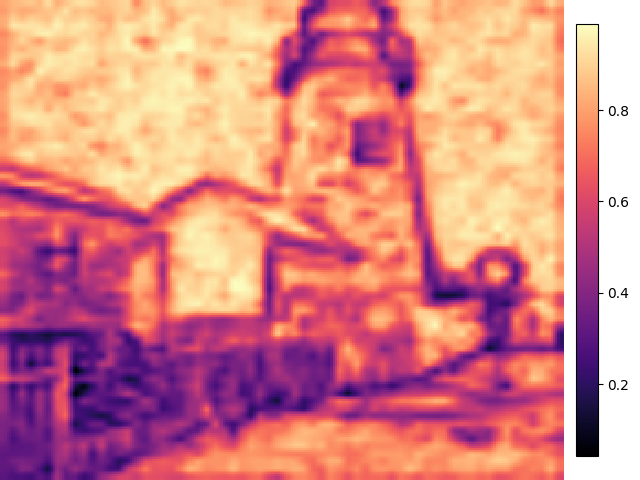}
    \end{minipage}
        \begin{minipage}[t]{0.32\linewidth}
        \centering
        \includegraphics[width=1\columnwidth,height=2.5cm]{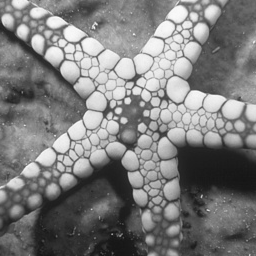}\\
        \includegraphics[width=1\columnwidth,height=2.5cm]{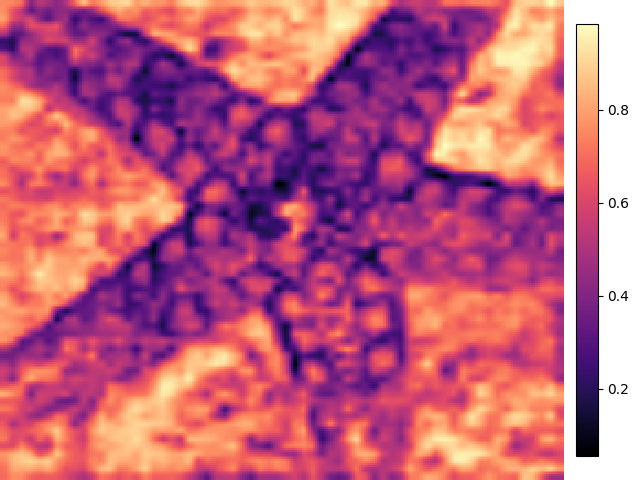}
    \end{minipage}
    \vspace{2pt}
    \caption{Visualization of the number of neighbors in different locations. The results are normalized and presented in the form of heat maps. The lighter color indicates a larger number of neighbors. One can see that the number of neighbors changes with the frequency of image content, demonstrating that our dynamic graph method can assign neighboring nodes according to demand.}
    \label{fig:ht_k}
    \vspace{-8pt}
\end{figure}

\textbf{Feature aggregation.} Guided by the adjacency matrix $\mathbf{A}$, the feature aggregation process is a weighted sum of all connected neighbors, which is represented as:
\begin{equation}
    \hat{\mathbf{p}}_{i} = \sum_{j\in \mathcal{N}_{i}}\alpha_{ij}\cdot \mathbf{p}_{j}^{\prime \prime} = \sum_{j\in \mathcal{N}_{i}}\alpha_{ij}\cdot f_{node}(\mathbf{p}_{j}).
    \label{sum_our}
\end{equation}
Then we extract all feature patches from the graph and utilize the fold operation to combine this array of updated local patches into a feature map, which can be viewed as the inverse of the unfold operation. Since there exist overlaps between feature patches, we use the average operation to deal with the overlapped areas. This strategy can also suppress the blocking effect in the final output. A global residual connection is employed in GFAM to further enhance the output. Thus, the output of GFAM is expressed as:
\begin{equation}
    \mathbf{F}_{out}=\mathbf{F}_{in}+Fold(\{\hat{\mathbf{p}}_{i}\}_{i=1}^{N}).
\end{equation}

To stabilize the training process of graph convolution, we extend our method to employ a multi-head graph to be beneﬁcial, represented as M-GFAM in Fig.~\ref{network}. The multi-heads design allows our method to jointly aggregate information from different representation subspaces at different positions. Speciﬁcally, $K$ independent heads execute the graph-based feature aggregation. Their results are concatenated together and once again projected by a $1\times 1$ convolutional layer ($f_{merge}$). Let us denote $\mathbf{F}_{out}^k$ as the output of the $k$-th head. The final output of M-GFAM can be calculated as $\mathbf{F}_{out}^{MH} = f_{merge}(\mathop {||}\limits_{k=1}^{K}\mathbf{F}_{out}^k)$.

\subsection{Analyze and Discussion}
As mentioned previously, our improved graph model can construct robust long-range correlations based on feature patches, and the number of neighbors dynamically changes with different nodes. In this subsection, we use the visualization results in Fig.~\ref{fig:ht_graph},~\ref{fig:ht_k} to demonstrate these merits.

\textbf{Robust long-range correlations.} In Fig.~\ref{fig:ht_graph}, we show the construction of graph connections of some query patches. The location of each query patch is labeled with a red box. For illustration purpose, we only present a limited number of neighbors (labeled with green boxes) with the highest attention weights. One can see that even the images are highly corrupted, our patch-wise graph method can still capture satisfied long-range correlations, and the adjacency matrix accurately filters out correlations with low importance.

\textbf{Dynamic graph connections.} Fig.~\ref{fig:ht_k} presents the normalized number of neighbors of each query patch at different locations. One can see that the number of neighbors follows distinct distributions over the frequency of image content, demonstrating that our dynamic graph method can adaptively select informative regions to aggregate the most useful information for each query patch.          
\begin{table*}[t]
\caption{Quantitative results (PSNR and SSIM) of gray-scale image denoising. Best and second-best results are \textbf{highlighted} and \underline{underlined}.}
\small
\centering
\begin{tabular}{c c c c c c c c c c}
\hline
Dataset & $\bm{\sigma}$ & BM3D \cite{BM3D} & DnCNN \cite{dncnn} & FFDNet \cite{ffdnet} & N3Net \cite{n3net} & NLRN \cite{nlrn} & GCDN \cite{gcdn} & DAGL (Ours)\\
\hline \hline
\multirow{4}*{Set12} & 15 & 32.37/0.8952 & 32.86/0.9031 & 32.75/0.9027 & 33.03/0.9056 & \underline{33.16}/0.9070 & 33.14/\underline{0.9072}& \textbf{33.28}/\textbf{0.9100} \\
& 25 & 29.96/0.8504 & 30.44/0.8622 & 30.43/0.8634 & 30.55/0.8648 & \underline{30.80}/\underline{0.8689} & 30.78/0.8687& \textbf{30.93}/\textbf{0.8720}\\
& 50 & 26.70/0.7676 & 27.19/0.7829 & 27.31/0.7903 & 27.43/0.7948 & \underline{27.64}/\underline{0.7980} & 27.60/0.7957& \textbf{27.81}/\textbf{0.8042}\\
\hline
\hline
\multirow{4}*{BSD68} & 15 & 31.07/0.8717 & 31.73/0.8907 & 31.63/0.8902 & 31.78/0.8927 & \underline{31.88}/0.8932 & 31.83/\underline{0.8933} & \textbf{31.93}/\textbf{0.8953}\\
& 25 & 28.57/0.8013 & 29.23/0.8278 & 29.19/0.8289  & 29.30/0.8321 & \underline{29.41}/0.8331 & 29.35/\underline{0.8332}& \textbf{29.46}/\textbf{0.8366}\\
& 50 & 25.62/0.6864 & 26.23/0.7189 & 26.29/0.7345 & 26.39/0.7293 & \underline{26.47}/0.7298 & 26.38/\textbf{0.7389}& \textbf{26.51}/\underline{0.7334}\\
\hline
\hline
\multirow{4}*{Urban100} & 15 & 32.35/0.9220 & 32.68/0.9255 & 32.43/0.9273 & 33.08/0.9333 & 33.45/0.9354 & \underline{33.47}/\underline{0.9358} & \textbf{33.79}/\textbf{0.9393}\\
& 25 & 29.71/0.8777 & 29.97/0.8797 & 29.92/0.8887 & 30.19/0.8925 & 30.94/0.9018 & \underline{30.95}/\underline{0.9020}& \textbf{31.39}/\textbf{0.9093}\\
& 50 & 25.95/0.7791 & 26.28/0.7874 & 26.52/0.8057 & 26.82/0.8184 & \underline{27.49}/\underline{0.8279} & 27.41/0.8160& \textbf{27.97}/\textbf{0.8423}\\
\hline
\hline
\multicolumn{2}{c}{Parameters$\downarrow$}& - &0.56M & 0.49M & 0.72M & 0.35M & 5.99M &5.62M\\
\hline
\end{tabular}
\label{tb_sdn}
\vspace{-5pt}
\end{table*}

\begin{figure*}[h]
\centering
\begin{minipage}{0.19\linewidth}
\centering
\footnotesize
\includegraphics[width=.95\columnwidth,height=2.9cm]{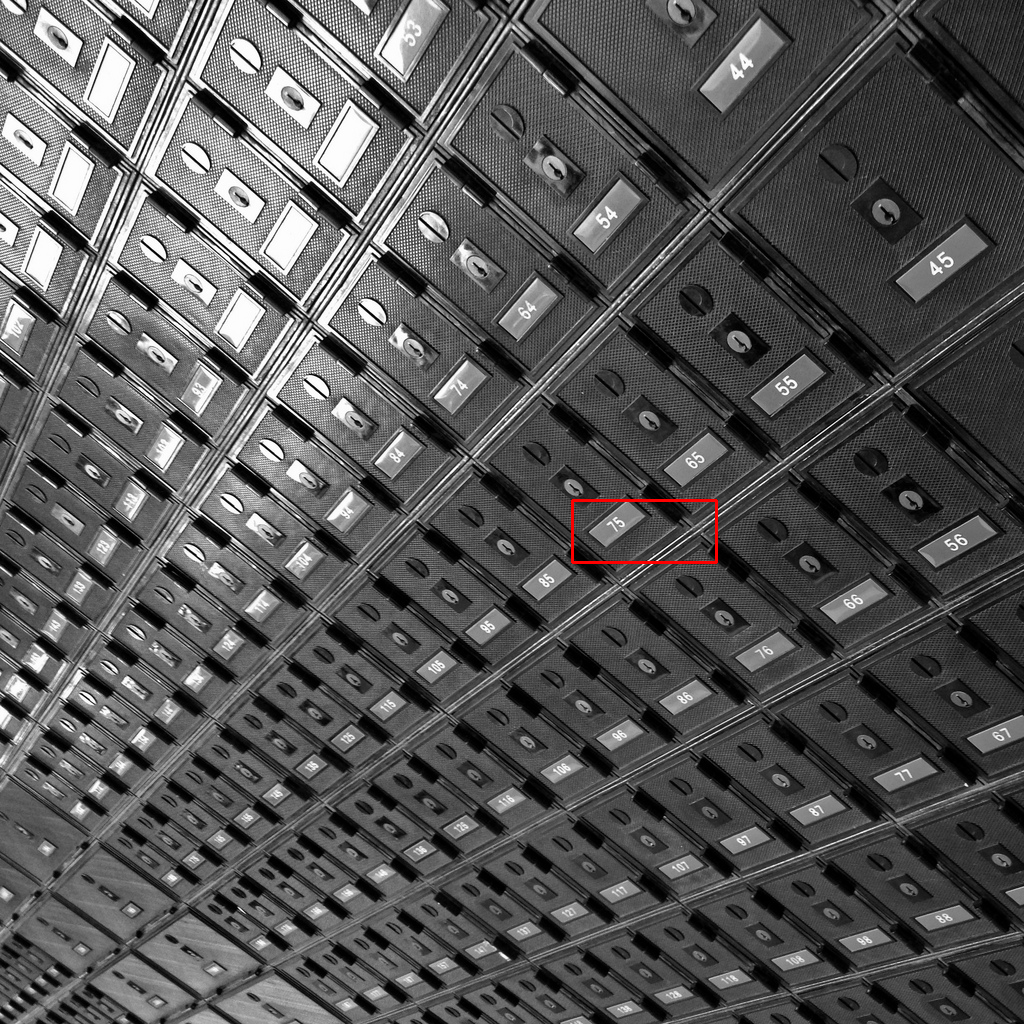}\\Urban100: img006
\end{minipage}
\begin{minipage}{0.19\linewidth}
\centering
\small
\includegraphics[width=.95\columnwidth,height=1.2cm]{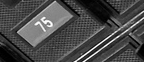}\\HQ (PSNR/SSIM)\\
\includegraphics[width=.95\columnwidth,height=1.2cm]{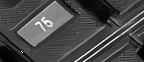}\\N3Net (27.40/0.8457)
\end{minipage}
\begin{minipage}{0.19\linewidth}
\centering
\small 
\includegraphics[width=.95\columnwidth,height=1.2cm]{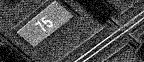}\\Noisy (20.60/0.5748)\\
\includegraphics[width=.95\columnwidth,height=1.2cm]{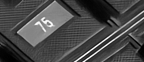}\\NLRN (28.46/0.8688)
\end{minipage}
\begin{minipage}{0.19\linewidth}
\centering
\small
\includegraphics[width=.95\columnwidth,height=1.2cm]{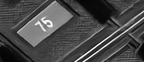}\\DnCNN (26.38/0.8052)\\
\includegraphics[width=.95\columnwidth,height=1.2cm]{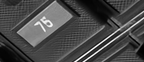}\\GCDN (28.70/0.8742)
\end{minipage}
\begin{minipage}{0.19\linewidth}
\centering
\small
\includegraphics[width=.95\columnwidth,height=1.2cm]{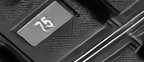}\\FFDNet (27.05/0.8359)\\
\includegraphics[width=.95\columnwidth,height=1.2cm]{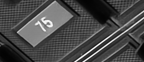}\\DAGL (29.29/0.8960)
\end{minipage}\\
\normalsize
\vspace{2pt}
\caption{Visual comparison of gray-scale image denoising of various methods on one sample from Urban100 with noise level $\sigma =25$.}
\label{im_urban}
\vspace{-8pt}
\end{figure*}

\section{Experiments}
To demonstrate the superiority of our proposed model, we apply our method to four typical image restoration tasks: synthetic image denoising, real image denoising, image demosaicing, and image compression artifacts reduction. For synthetic image denoising, image demosaicing, and image compression artifacts reduction tasks, we train our DAGL on DIV2K~\cite{div2k} dataset, which contains 800 high-quality images. For real image denoising, we apply the commonly used SIDD~\cite{sidd} dataset as the training data, which contains 160 images corrupted by realistic noise and corresponding high-quality labels. In each task, we utilize commonly used test sets for evaluation and report PSNR and SSIM \cite{ssim} to evaluate the performance of each method. Our model is trained on an Nvidia Tesla V100 GPU with the initial learning rate $lr = 1\times10^{-4}$ and performs halving per 200 epochs. During training, we employ Adam optimizer, and each mini-batch contains 32 images with the size of $64\times64$ randomly cropped from training data. The training process can be completed within two days.

\subsection{Synthetic Image Denoising} 
We compare our proposed model with some state-of-the-art denoising methods, including some well-known denoisers, \textit{e.g.}, BM3D \cite{BM3D}, DnCNN \cite{dncnn}, and FFDNet \cite{ffdnet}, and recent competitive non-local denoisers such as N3Net \cite{n3net} and NLRN \cite{nlrn}. Furthermore, we also compare our method with the graph-based denoiser: GCDN \cite{gcdn}. The standard test sets: Urban100 \cite{urban100}, BSD68 \cite{bsd68}, and Set12 are applied to evaluate each method. Additive white Gaussian noise (AWGN) with different noise levels (15, 25, 50) are added to the clean images. The quantitative results (PSNR and SSIM) and the number of parameters of different methods are shown in Table~\ref{tb_sdn}. The visual comparison is shown in Fig.~\ref{im_urban}. One can see that our method has the best performance in all noise levels and produces higher visual quality while maintaining a moderate number of parameters. 

\subsection{Real Image Denoising}
\begin{figure*}[h]
\centering
\small 
\begin{minipage}[t]{0.118\linewidth}
\centering
\includegraphics[width=1\columnwidth]{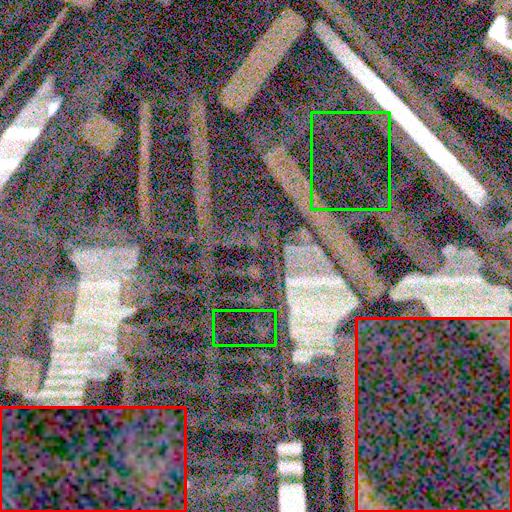}\\Noisy\\\vspace{3pt}
\includegraphics[width=1\columnwidth]{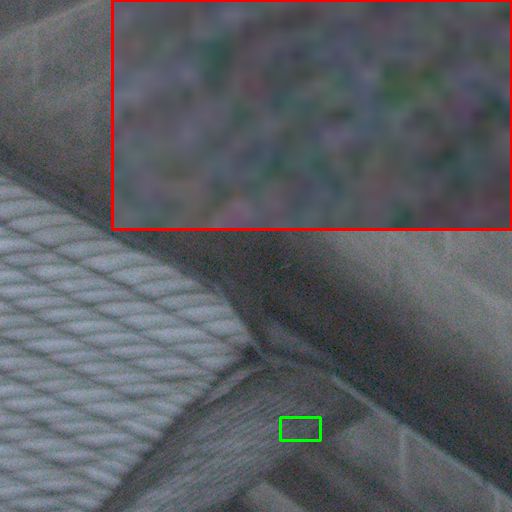}\\Noisy
\end{minipage}
\begin{minipage}[t]{0.118\linewidth}
\centering
\includegraphics[width=1\columnwidth]{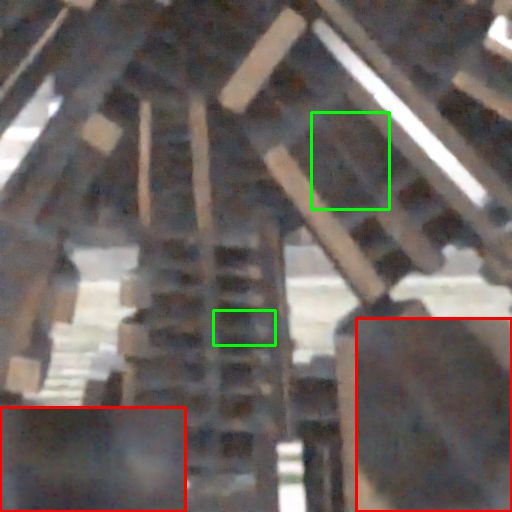}\\CBDNet \cite{cbdnet}\\\vspace{3pt}
\includegraphics[width=1\columnwidth]{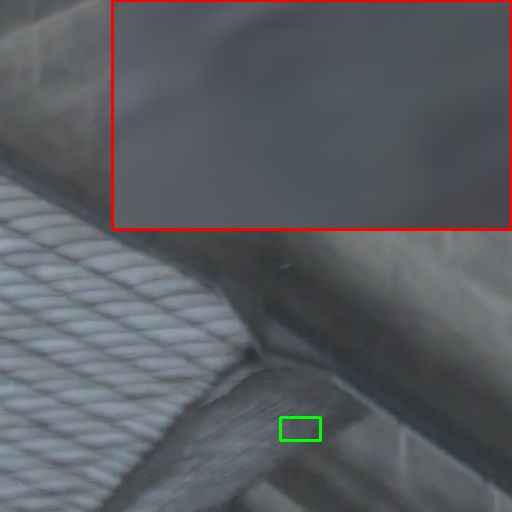}\\CBDNet \cite{cbdnet}
\end{minipage}
\begin{minipage}[t]{0.118\linewidth}
\centering
\includegraphics[width=1\columnwidth]{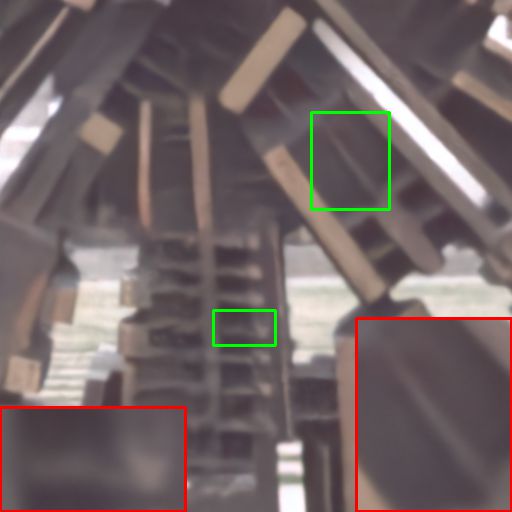}\\VDNet \cite{vdnet}\\\vspace{3pt}
\includegraphics[width=1\columnwidth]{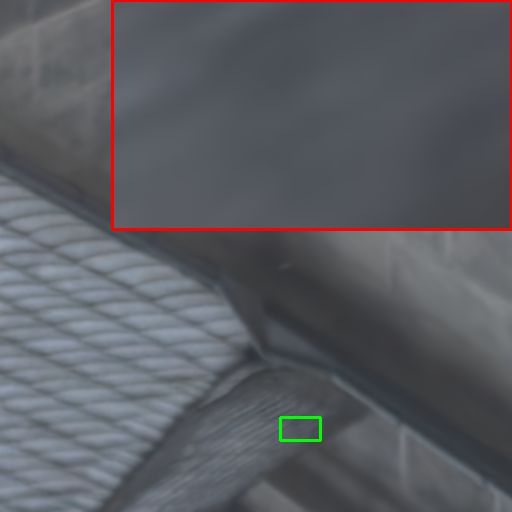}\\VDNet \cite{vdnet}
\end{minipage}
\begin{minipage}[t]{0.118\linewidth}
\centering
\includegraphics[width=1\columnwidth]{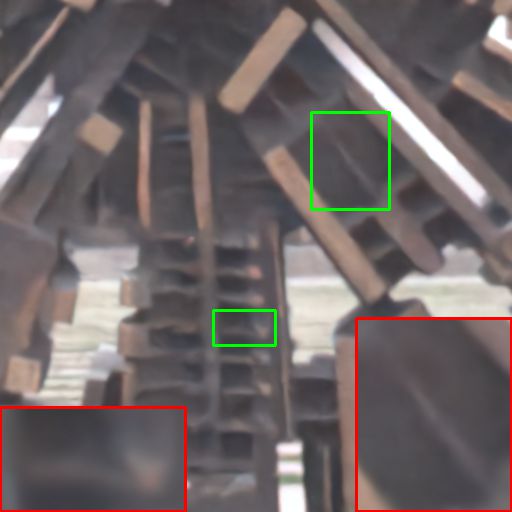}\\CycleISP \cite{cycleisp}\\\vspace{3pt}
\includegraphics[width=1\columnwidth]{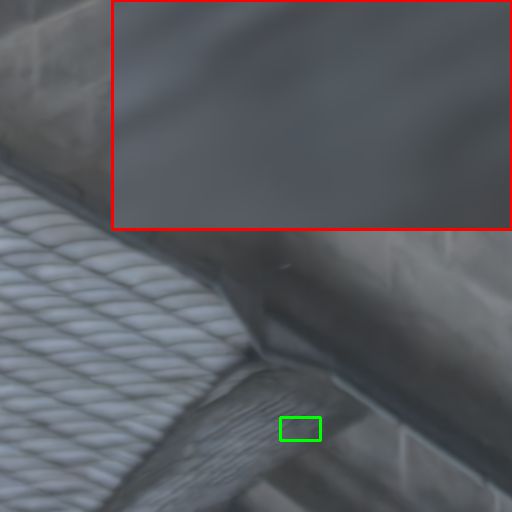}\\CycleISP \cite{cycleisp}
\end{minipage}
\begin{minipage}[t]{0.118\linewidth}
\centering
\includegraphics[width=1\columnwidth]{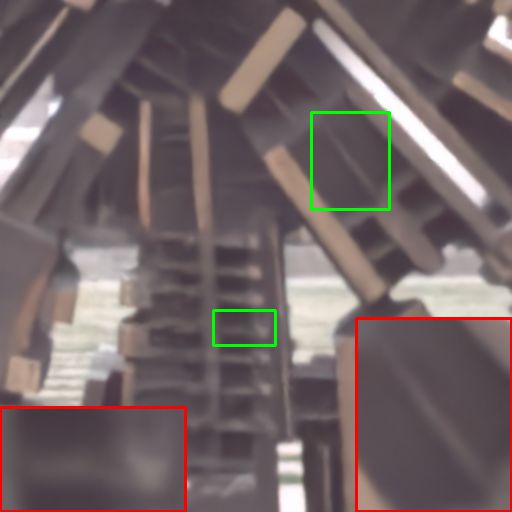}\\GDANet \cite{gdanet}\\\vspace{3pt}
\includegraphics[width=1\columnwidth]{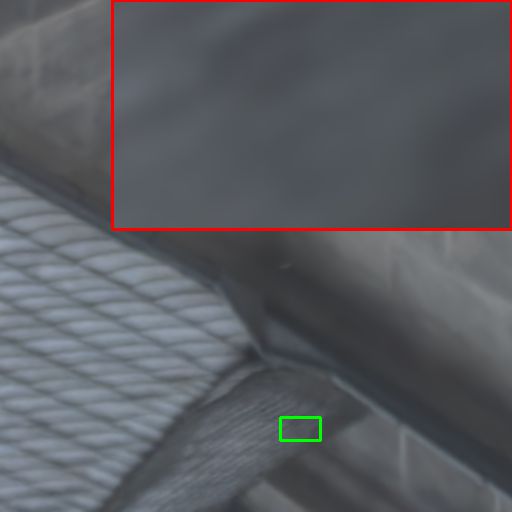}\\GDANet \cite{gdanet}
\end{minipage}
\begin{minipage}[t]{0.118\linewidth}
\centering
\includegraphics[width=1\columnwidth]{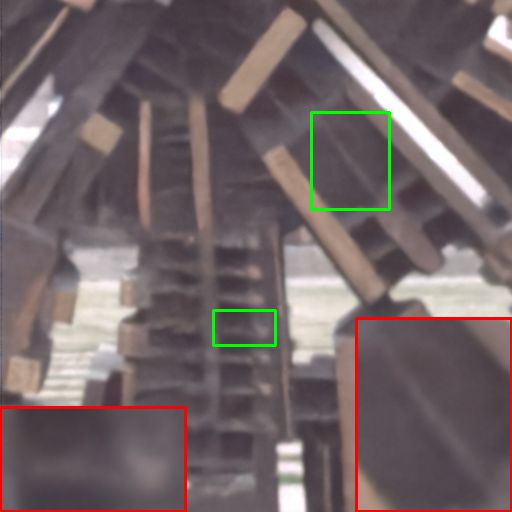}\\AINDNet \cite{aindnet}\\\vspace{3pt}
\includegraphics[width=1\columnwidth]{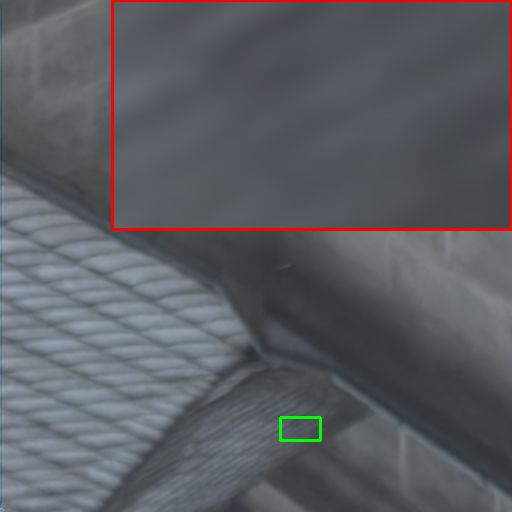}\\AINDNet \cite{aindnet}
\end{minipage}
\begin{minipage}[t]{0.118\linewidth}
\centering
\includegraphics[width=1\columnwidth]{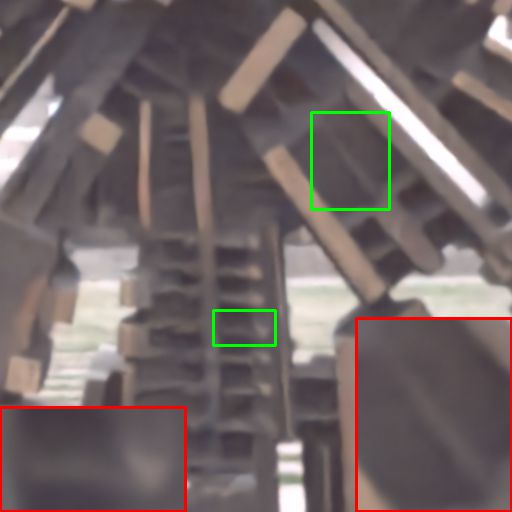}\\DeamNet \cite{deamnet}\\\vspace{3pt}
\includegraphics[width=1\columnwidth]{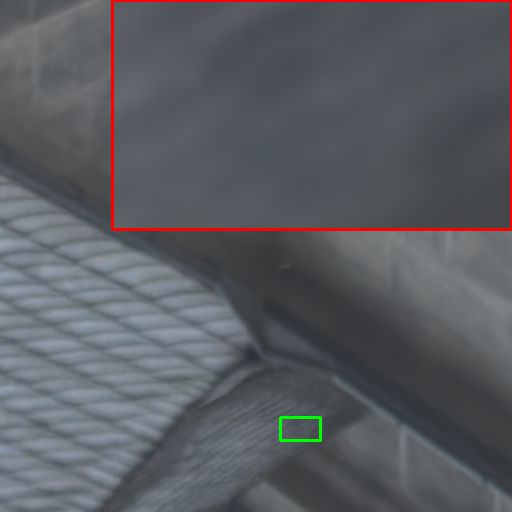}\\DeamNet \cite{deamnet}
\end{minipage}
\begin{minipage}[t]{0.118\linewidth}
\centering
\includegraphics[width=1\columnwidth]{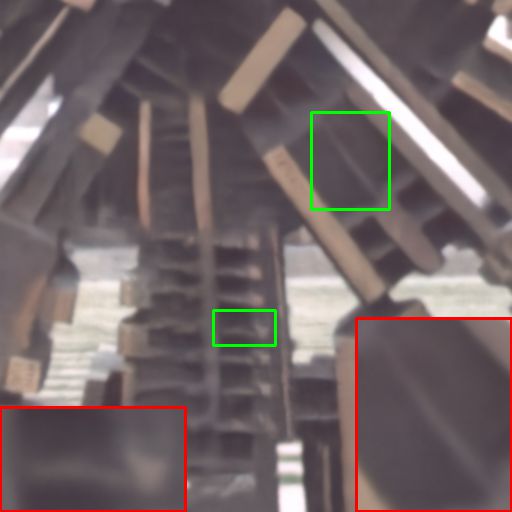}\\DAGL (Ours)\\\vspace{4pt}
\includegraphics[width=1\columnwidth]{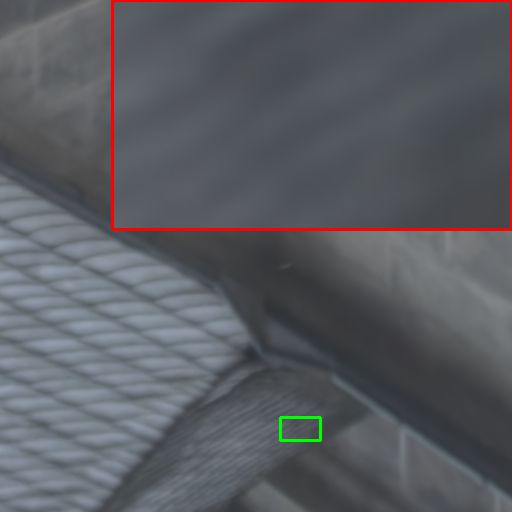}\\DAGL (Ours)
\end{minipage}
\centering
\caption{Visual comparison of real image denoising application of various methods. These noisy images come from DND \cite{dnd} dataset.}
\label{im_rdn} 
\end{figure*}

To further demonstrate the merits of our proposed method, we apply it to the more challenging task of real image denoising. Unlike synthetic image denoising, in this case, images are corrupted by realistic noise with unknown distribution and noise levels. We compare our method with some competitive denoisers \cite{BM3D,dncnn,ffdnet,twsc,cbdnet} and some very recent methods \cite{ridnet,vdnet,cycleisp,gdanet,aindnet,deamnet}. The commonly used DND \cite{dnd} dataset is utilized for evaluation. Note that the high-quality labels of DND are not available. We get the evaluation results from the official website. The quantitative results are shown in Table~\ref{tb_rdn}, and we further provide a visual comparison of different methods in Fig.~\ref{im_rdn}. In comparison, our algorithm recovers the actual texture and structures without compromising on the removal of noise, and our method has good robustness to both high-intensity and low-intensity noise. Even compared with top-performing methods \cite{cycleisp,gdanet,aindnet,deamnet}, our proposed DAGL can achieve better performance with attractive model parameters.

\begin{table}[h]
\caption{The quantitative results of real image denoising on DND dataset~\cite{dnd}.}
\small
\centering
\begin{tabular}{c c c c c}
\hline
\multirow{2}*{Algorithm}&\multirow{2}*{Params$\downarrow$}&\multirow{2}*{Mode}&\multicolumn{2}{c}{sRGB}\\
\cline{4-5}
&&& PSNR$\uparrow$ & SSIM$\uparrow$ \\
\hline
\hline
BM3D \cite{BM3D}&-& Non-blind & 34.51& 0.851 \\
CDnCNN \cite{dncnn}&0.67M& Blind & 32.43 & 0.790 \\
CFFDNet \cite{ffdnet}&0.85M &Non-blind & 37.61 & 0.914\\
TWSC \cite{twsc}&- & Blind& 37.94 & 0.940 \\
CBDNet \cite{cbdnet}&4.36M &Blind & 38.06 & 0.942\\
RIDNet \cite{ridnet}& 1.50M&Blind & 39.23 & 0.952\\
VDNet \cite{vdnet} & 7.82M & Blind & 39.38 & 0.952\\
CycleISP \cite{cycleisp} & 2.60M & Blind & 39.56 & \underline{0.956}\\
GDANet \cite{gdanet} & 9.15M & Blind & 39.58 & 0.954\\
AINDNet \cite{aindnet} & 13.76M &Blind & 39.37 & 0.951 \\
DeamNet \cite{deamnet} & 2.25M & Blind & \underline{39.70} & 0.953\\
DAGL (Ours) & 5.62M & Blind & \textbf{39.83} & \textbf{0.957}\\
\hline
\end{tabular}
\label{tb_rdn}
\vspace{-8pt}
\end{table}

\begin{table*}[t]
\caption{Quantitative results of image compression artifact reduction. Best and second-best results are \textbf{highlighted} and \underline{underlined}.}
\footnotesize
\begin{center}
\begin{tabular}{c c c c c c c c c c}
\hline 
Dataset &$q$ & JPEG & SA-DCT \cite{sadct} & ARCNN \cite{arcnn} & TNRD \cite{tnrd} & DnCNN \cite{dncnn} & RNAN \cite{rnan} & DUN~\cite{dun} & DAGL (ours)\\
\hline \hline
\multirow{4}*{LIVE1} & 10 & 27.77/0.7905 & 28.86/0.8093 & 28.98/0.8076 & 29.15/0.8111 & 29.19 /0.8123& \underline{29.63}/\underline{0.8239} & 29.61/0.8232 & \textbf{29.70}/\textbf{0.8245}\\
& 20 & 30.07/0.8683 & 30.81/0.8781 & 31.29/0.8733 & 31.46/0.8769 & 31.59/0.8802 & \underline{32.03}/\underline{0.8877} & 31.98/0.8869 & \textbf{32.12}/\textbf{0.8887} \\
& 30 & 31.41/0.9000 & 32.08/0.9078 & 32.69/0.9043 & 32.84/0.9059 & 32.98/0.9090 & \underline{33.45}/\underline{0.9149} & 33.38/0.9142 & \textbf{33.54}/\textbf{0.9156} \\
& 40 & 32.35/0.9173 & 32.99/0.9240 & 33.63/0.9198 & -/- & 33.96/0.9247 & \underline{34.47}/\underline{0.9299} & 34.32/0.9289 & \textbf{34.53}/\textbf{0.9305}\\
\hline
\hline
\multirow{4}*{Classic5} & 10 & 27.82/0.7800 & 28.88/0.8071 & 29.04/0.7929 & 29.28/0.7992 & 29.40/0.8026 & \underline{29.96}/\underline{0.8178} & 29.95/0.8171 & \textbf{30.08}/\textbf{0.8196}\\
& 20 & 30.12/0.8541 & 30.92/0.8663 & 31.16/0.8517 & 31.47/0.8576 & 31.63/0.8610 & 32.11/\underline{0.8693} & \underline{32.11}/0.8689 & \textbf{32.35}/\textbf{0.8719} \\
& 30 & 31.48/0.8844 & 32.14/0.8914 & 32.52/0.8806 & 32.74/0.8837 & 32.91/0.8861 & \underline{33.38}/\underline{0.8924} & 33.33/0.8916 & \textbf{33.59}/\textbf{0.8942}\\
& 40 & 32.43/0.9011 & 33.00/0.9055 & 33.34/0.8953 & -/- & 33.77/0.9003 & \underline{34.27}/\underline{0.9061} & 34.10/0.9045 & \textbf{34.41}/\textbf{0.9069}\\
\hline
\hline
\multicolumn{2}{c}{Parameters}&-&-&0.12M&-&0.56M&8.96M & 10.5M &5.62M\\
\hline
\end{tabular}
\end{center}
\label{tb_db}
\vspace{-8pt}
\end{table*}

\begin{figure*}[h]
\centering
\small
\begin{minipage}[t]{0.135\linewidth}
\centering
\includegraphics[width=1\columnwidth,height=2.2cm]{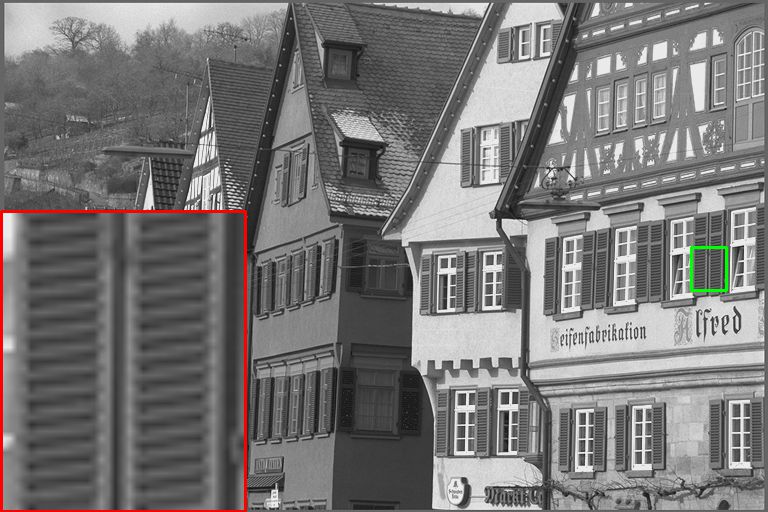}\\HQ\\PSNR/SSIM\\\vspace{3pt}
\includegraphics[width=1\columnwidth,height=2.2cm]{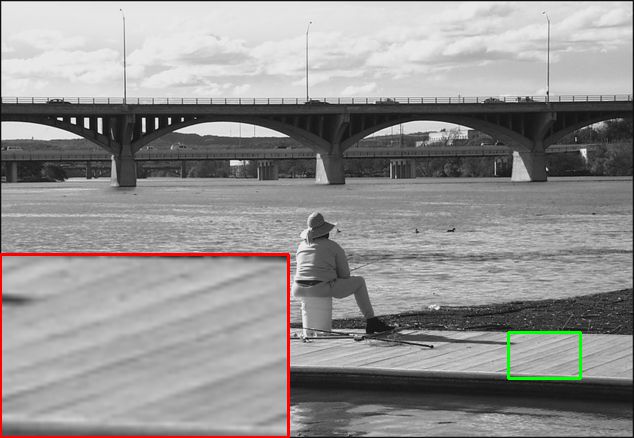}\\HQ\\PSNR/SSIM
\end{minipage}%
\hspace{0.01mm}
\begin{minipage}[t]{0.135\linewidth}
\centering
\includegraphics[width=1\columnwidth,height=2.2cm]{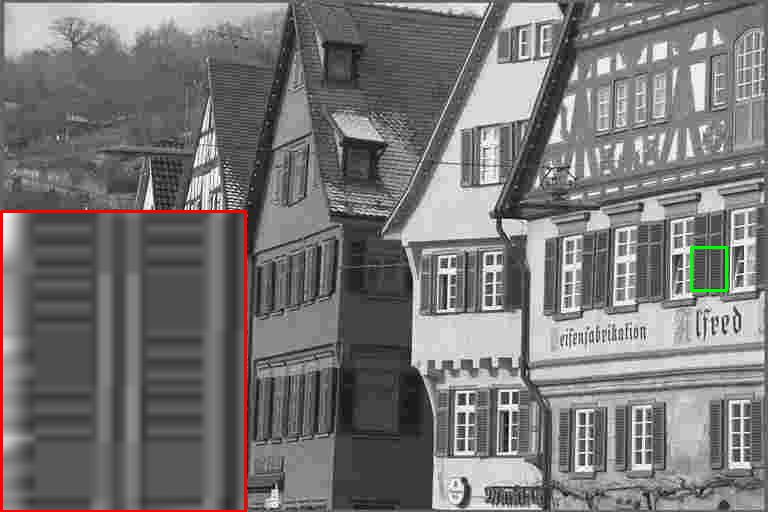}\\JPEG (q=10)\\25.07/0.7632\\\vspace{3pt}
\includegraphics[width=1\columnwidth,height=2.2cm]{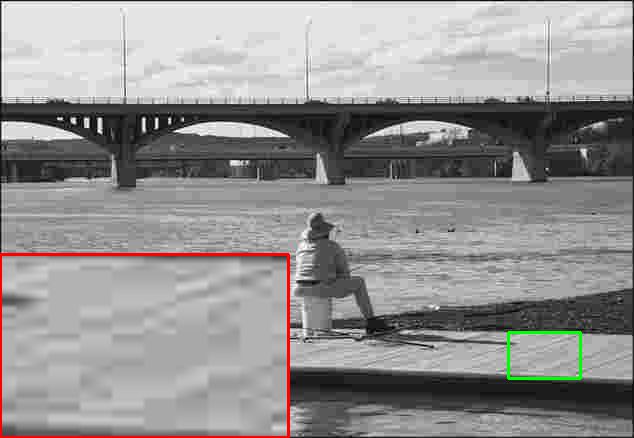}\\JPEG (q=10)\\27.59/0.7747
\end{minipage}%
\hspace{0.01mm}
\begin{minipage}[t]{0.135\linewidth}
\centering
\includegraphics[width=1\columnwidth,height=2.2cm]{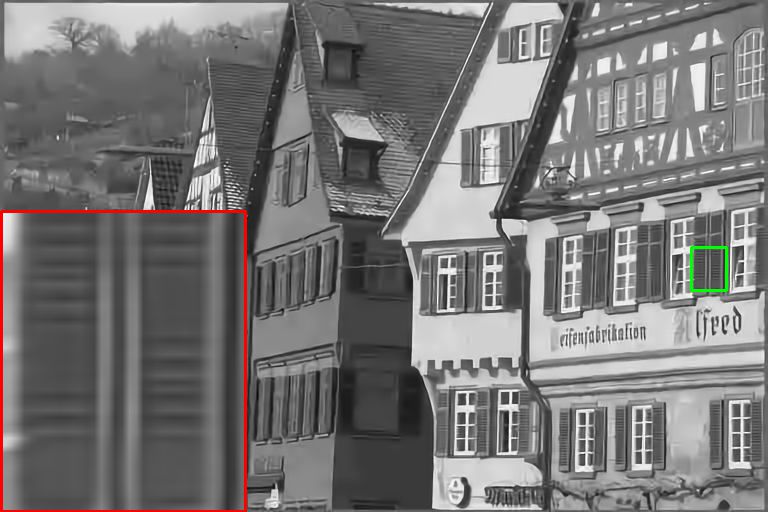}\\TNRD \cite{tnrd}\\26.64/0.8055\\\vspace{3pt}
\includegraphics[width=1\columnwidth,height=2.2cm]{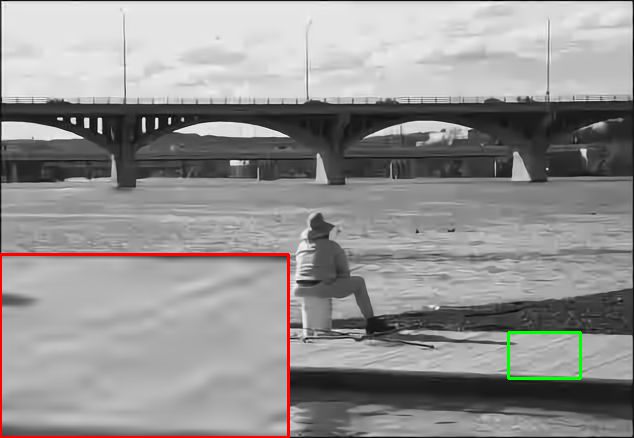}\\TNRD \cite{tnrd}\\28.94/0.8111
\end{minipage}
\begin{minipage}[t]{0.135\linewidth}
\centering
\includegraphics[width=1\columnwidth,height=2.2cm]{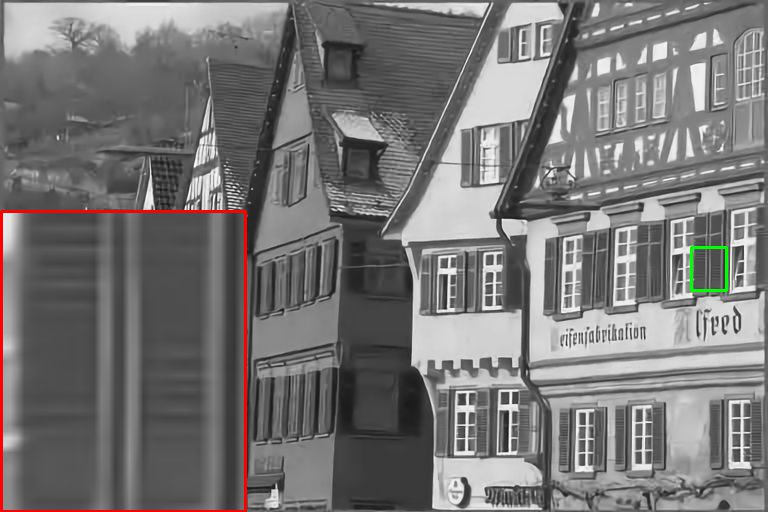}\\DnCNN \cite{dncnn}\\26.75/0.8066\\\vspace{3pt}
\includegraphics[width=1\columnwidth,height=2.2cm]{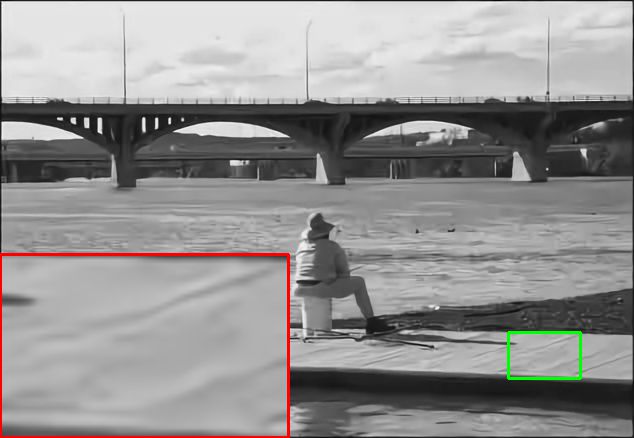}\\DnCNN \cite{dncnn}\\28.96/0.8122
\end{minipage}
\begin{minipage}[t]{0.135\linewidth}
\centering
\includegraphics[width=1\columnwidth,height=2.2cm]{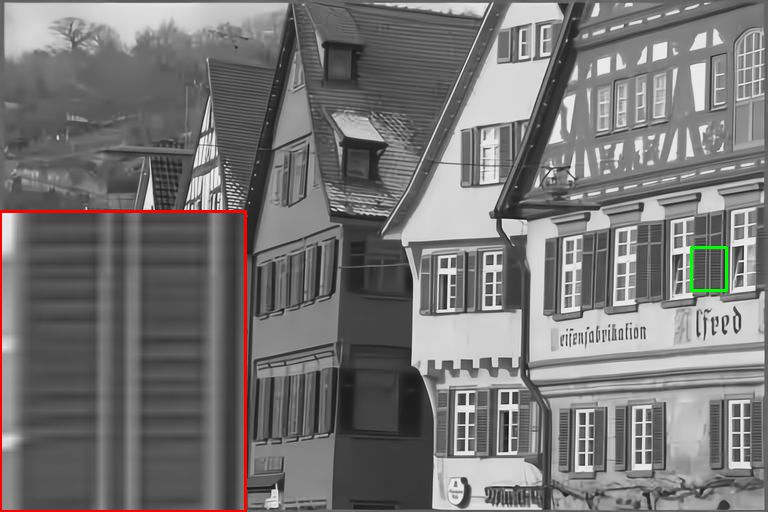}\\RNAN \cite{rnan}\\27.58/0.8314\\\vspace{3pt}
\includegraphics[width=1\columnwidth,height=2.2cm]{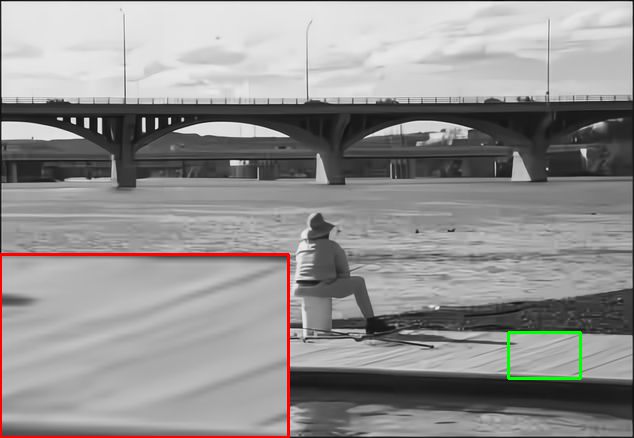}\\RNAN \cite{rnan}\\29.49/0.8305
\end{minipage}
\begin{minipage}[t]{0.135\linewidth}
\centering
\includegraphics[width=1\columnwidth,height=2.2cm]{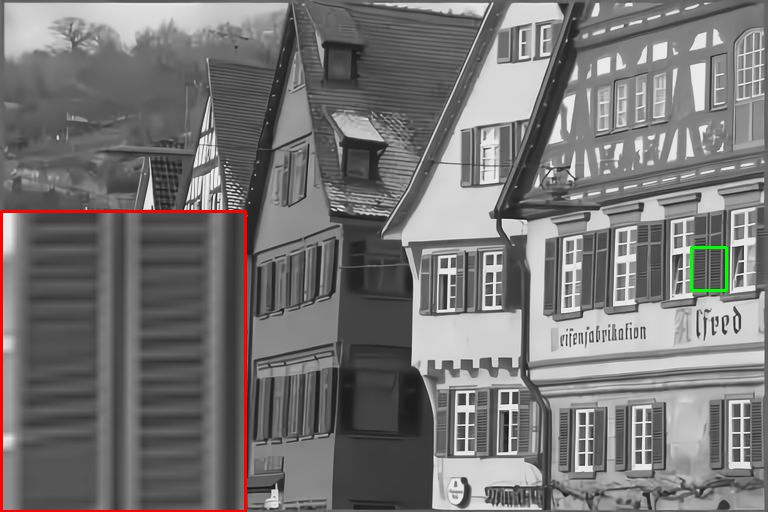}\\DUN \cite{dun}\\27.67/0.8320\\\vspace{3pt}
\includegraphics[width=1\columnwidth,height=2.2cm]{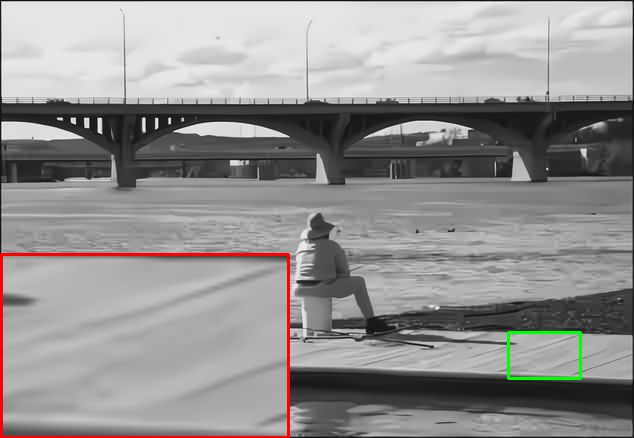}\\DUN \cite{dun}\\29.50/0.8301
\end{minipage}
\begin{minipage}[t]{0.135\linewidth}
\centering
\includegraphics[width=1\columnwidth,height=2.2cm]{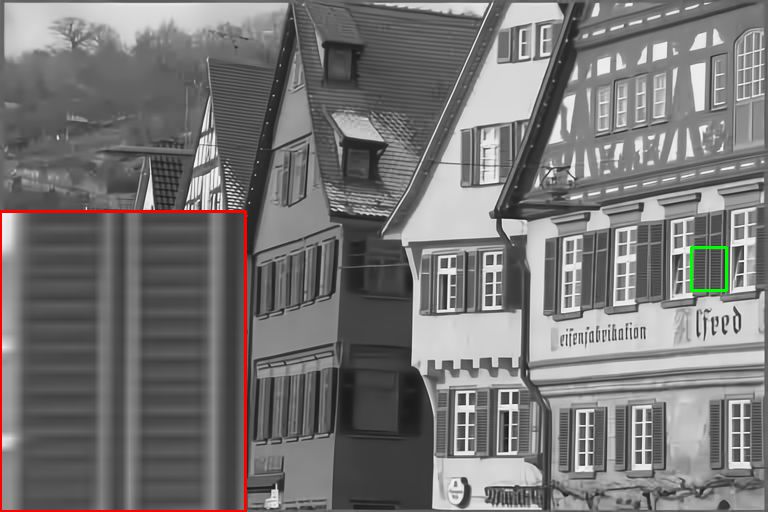}\\DAGL (Ours)\\27.82/0.8379\\\vspace{3pt}
\includegraphics[width=1\columnwidth,height=2.2cm]{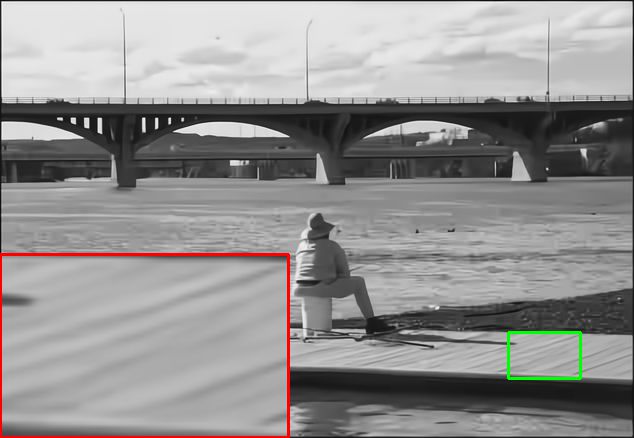}\\DAGL (Ours)\\29.53/0.8316
\end{minipage}
\centering
\small
\caption{Visual comparison of image compression artifact reduction application of various methods with JPEG quality q = 10.}
\label{im_car} 
\end{figure*}

\subsection{Image Compression Artifact Reduction} 
For this application, we compare our DAGL with some classic methods (\textit{e.g.}, SA-DCT \cite{sadct}, ARCNN \cite{arcnn}, TNRD~\cite{tnrd}) and recent competitive deep-learning methods (\textit{e.g.}, DnCNN \cite{dncnn}, RNAN \cite{rnan}, DUN~\cite{dun}). To demonstrate the superiority of our DAGL, we apply the same setting as DnCNN and DUN, \textit{i.e.}, using a single model to handle all degradation levels. The compressed images are generated by Matlab standard JPEG encoder with quality factors $q\in \{10, 20, 30, 40\}$. We evaluate the performance of each method on the commonly used Classic5 \cite{sadct} and LIVE1 \cite{live1} test sets. The quantitative results are presented in Table~\ref{tb_db}. One can see that under the evaluation of both PSNR and SSIM, our proposed method achieves the best performance on all test sets and quality factors with a single model. In addition, the number of parameters of our DAGL is much fewer than the top-performing methods \cite{rnan,dun}. The visual comparison is shown in Fig.~\ref{im_car}, presenting the better restoration quality of our proposed DAGL. 


\begin{table}[t]
\caption{Quantitative comparison of image demosaicing.}
\footnotesize
\centering
\begin{tabular}{c c c c c}
\hline
\multirow{2}{*}{Method}& \multirow{2}{*}{Params} & \multicolumn{3}{c}{PSNR/SSIM} \\ 
\cline{3-5} & & McMaster18  & Kodak24  & Urban100\\
\hline
\hline
Mosaic     & -          & 9.17/0.1674       & 8.56/0.0682 & 7.48/0.1195       \\ 
IRCNN & 0.19M     & 37.47/0.9615       & 40.41/0.9807  & 36.64/0.9743      \\ 
RNAN  & 8.96M     & \underline{39.71}/\underline{0.9725}   & \underline{43.09}/\underline{0.9902}    & \underline{39.75}/\underline{0.9848}       \\ 
DAGL         & 5.62M           &\textbf{39.84}/\textbf{0.9735} &	\textbf{43.21}/\textbf{0.9910}	& \textbf{40.20}/\textbf{0.9854}       \\ 
\hline
\end{tabular}
\label{t_dmsk}
\vspace{-8pt}
\end{table}

\begin{figure*}[h]
\centering
\small 
\centering
\begin{minipage}[t]{0.16\linewidth}
\centering
\includegraphics[width=.96\columnwidth,height=2.5cm]{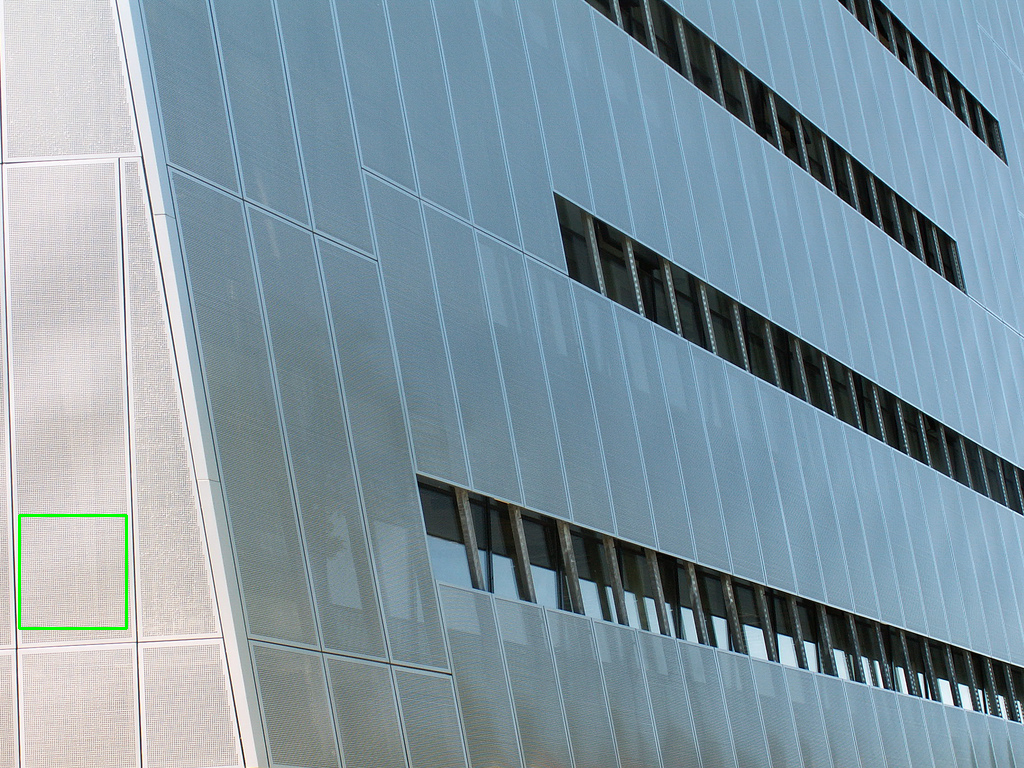}\\Original image\\Urban100: img026
\end{minipage}
\begin{minipage}[t]{0.16\linewidth}
\centering
\includegraphics[width=.96\columnwidth,height=2.5cm]{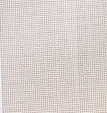}\\HQ\\PSNR/SSIM
\end{minipage}
\centering
\begin{minipage}[t]{0.16\linewidth}
\centering
\includegraphics[width=.96\columnwidth,height=2.5cm]{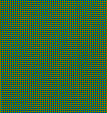}\\Mosaiced\\5.98/0.0395
\end{minipage}%
\hspace{0.25mm}
\centering
\begin{minipage}[t]{0.16\linewidth}
\centering
\includegraphics[width=.96\columnwidth,height=2.5cm]{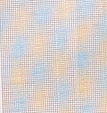}\\IRCNN \cite{ircnn}\\33.53/0.9235
\end{minipage}
\centering
\begin{minipage}[t]{0.16\linewidth}
\centering
\includegraphics[width=.96\columnwidth,height=2.5cm]{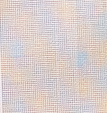}\\RNAN \cite{rnan}\\35.79/0.9519
\end{minipage}
\centering
\begin{minipage}[t]{0.16\linewidth}
\centering
\includegraphics[width=.96\columnwidth,height=2.5cm]{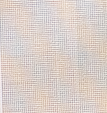}\\DAGL\\36.35/0.9536
\end{minipage}
\centering
\vspace{2pt}
\caption{Visual comparison of image demosaicing. Following \cite{ircnn}, the corrupted images are generated via Matlab with Bayer mosaic.}
\label{im_dmsk}
\end{figure*}

\subsection{Image Demosaicing} 
In this task, we compare our method with RNAN~\cite{rnan} and IRCNN~\cite{ircnn} on McMaster18~\cite{ircnn}, Kodak24, and Urban100 test sets. The quantitative result is shown in Table~\ref{t_dmsk}, and the visual comparison is shown in Fig.~\ref{im_dmsk}. One can see that the degraded images have very low quality on both subjective and objective evaluations. IRCNN and RNAN can restore the low-quality images with a good result, but our method can still make an improvement. 

\subsection{Ablation Study}
In this subsection, we show the ablation study in Table \ref{tab:abs_1} and Table \ref{tab:abs_2} to investigate the effect of different components in our proposed DAGL. The experiment of ablation study is conducted on denoising task and evaluated on Urban100 test set. The noise level is set as $25$. In Table \ref{tab:abs_1}, we compare the performance of the variants of our proposed DAGL. In Table \ref{tab:abs_2}, we explore the gains brought by the number of graph-based feature aggregation modules (GFAMs) in depth (number of stages) and width (number of heads). 

\textbf{Patch-wise non-local correlation.} The non-local correlations in our DAGL are constructed based on feature patches instead of pixels. To study the effectiveness of this design, we compare our method with the commonly used non-local neural networks \cite{nonlocalnet}. Correctly, we replace the graph modules in our DAGL with non-local neural networks with one head (NL) and multiple heads (MHNL). The results are presented in Table \ref{tab:abs_1}. One can see that our patch-wise non-local method obviously outperforms the commonly used pixel-wise non-local method \cite{nonlocalnet}.

\textbf{Graph attention mechanism.} In this paper, we extend the graph attention mechanism to image restoration tasks. To demonstrate the effectiveness of this strategy, we replace the attention-weighted aggregation process with directly averaging, denoted as (w/o GAT) in Table \ref{tab:abs_1}. The performance reduction demonstrates the positive effect of the graph attention mechanism used in our DAGL.

\textbf{Dynamic graph connections.} Different from existing non-local image restoration methods, in our DAGL, the number of neighbors of each query patch is dynamic and adaptive. To demonstrate the effectiveness of this design, we remove the dynamic KNN module from our GFAM, leading to a fully connected non-local attention operation with a fixed number of neighbors for each query patch. This variant is denoted as (w/o THD) in Table \ref{tab:abs_1}. There are 0.11dB gains by using the dynamic KNN module, demonstrating the necessity of the dynamics in our graph model.

\begin{table}[h]
    \centering
        \caption{Evaluation results about variants of DAGL on Urban100 test set ($\sigma=25$). NL and MHNL represent replacing our M-GFAM with non-local neural networks and multi-heads non-local neural networks, respectively. (w/o THD) and (w/o GAT) refer to removing dynamic KNN module and removing graph attention mechanism, respectively.}
\small
    \begin{tabular}{c c c c c c}
    \hline
        Mode & NL & MHNL & w/o THD & w/o GAT & DAGL\\
    \hline
    \hline
         PSNR & 30.73 & 30.92 & 31.28 & 30.77 & 31.39\\
    \hline
    \end{tabular}
    \label{tab:abs_1}
\end{table}

\textbf{Block number.} In this part, we explore the gains brought by the number of graph-based feature aggregation modules (GFAM) in depth (number of stages) and width (number of heads). The results are shown in Table~\ref{tab:abs_2}. Note that case 1 is constructed by removing all GFAM from our DAGL, resulting in a simple ResNet. From the results, we can find that our proposed GFAM can significantly boost the image restoration performance, and the performance increases with the number of heads and stages. By making a trade-off between performance and computational complexity, we employ four heads and three stages in our proposed DAGL. 

\begin{table}[h]
    \centering
        \caption{Evaluation results on Urban100 ($\sigma=25$) test set of our proposed model with different numbers of graph-based feature aggregation modules (GFAMs).}
\small
    \begin{tabular}{c c c c}
    \hline
    Case Index & Number of heads-stages & Params  & PSNR\\
    \hline
    \hline
        1 & 0-0 & 1.23M &30.43\\
        2 & 1-3 & 2.79M &31.21\\
        3 & 2-3 & 3.77M &31.29\\
        4 & 3-3 & 4.74M &31.33\\
        5 (DAGL) & 4-3 & 5.62M &31.39\\
        6 & 5-3 & 6.71M &31.41\\
        7 & 4-4 & 7.22M &31.42\\
        8 & 4-2 & 4.12M &31.28\\
        9 & 4-1 & 2.51M & 31.05\\
    \hline
    \end{tabular}
    \label{tab:abs_2}
\end{table}

\section{Conclusion}
In this paper, we propose an improved graph attention model for image restoration. Unlike previous non-local image restoration methods, our model can assign an adaptive number of neighbors for each query item and construct long-range correlations based on feature patches. Furthermore, our proposed dynamic attentive graph learning can be easily extended to other computer vision tasks. Extensive experiments demonstrate that our proposed model achieves state-of-the-art performance on wide image restoration tasks: synthetic image denoising, real image denoising, image demosaicing, and compression artifact reduction. 

{\small
\bibliographystyle{ieee_fullname}
\bibliography{egbib}
}

\end{document}